\documentclass[10pt,twocolumn,letterpaper]{article}

\usepackage{iccv}
\usepackage{times}
\usepackage{epsfig}
\usepackage{graphicx}
\usepackage{amsmath}
\usepackage{amssymb}
\usepackage{booktabs}
\usepackage{multirow}
\usepackage[table]{xcolor}

\usepackage{pifont}
\usepackage{float}
\usepackage{bm}
\usepackage{enumitem}
\usepackage{capt-of,etoolbox}
\usepackage{mathtools}
\usepackage{bbm}

\newcommand{\e}[1]{{\small $#1$}}

\newcommand\blfootnote[1]{%
  \begingroup
  \renewcommand\thefootnote{}\footnote{#1}%
  \addtocounter{footnote}{-1}%
  \endgroup
}

\newcounter{alphasect}
\def\alphainsection{0}

\let\oldsection=\section
\def\section{%
  \ifnum\alphainsection=1%
    \addtocounter{alphasect}{1}
  \fi%
\oldsection}%

\renewcommand\thesection{%
  \ifnum\alphainsection=1%
    \Alph{alphasect}%
  \else%
    \arabic{section}%
  \fi%
}%

\newenvironment{alphasection}{%
  \ifnum\alphainsection=1%
    \errhelp={Let other blocks end at the beginning of the next block.}
    \errmessage{Nested Alpha section not allowed}
  \fi%
  \setcounter{alphasect}{0}
  \def\alphainsection{1}
}{%
  \setcounter{alphasect}{0}
  \def\alphainsection{0}
}%

\usepackage[pagebackref=true,breaklinks=true,letterpaper=true,colorlinks,bookmarks=false]{hyperref}

\iccvfinalcopy %

\ificcvfinal\fi

\begin{document}

\title{Harnessing the Spatial-Temporal Attention of Diffusion Models for High-Fidelity Text-to-Image Synthesis}

\author{
\textbf{Qiucheng Wu}${^{1*}}$, \textbf{Yujian Liu}$^{1*}$,
\textbf{Handong Zhao}${^2}$,\textbf{Trung Bui}${^2}$, \textbf{Zhe Lin}${^2}$, \textbf{Yang Zhang}${^3}$, \textbf{Shiyu Chang}${^1}$\\
$^1$UC Santa Barbara, $^2$Adobe Research, $^3$ MIT-IBM Watson AI Lab
\\\small\texttt{\{qiucheng, yujianliu\}@ucsb.edu}}

\let\oldtwocolumn\twocolumn
\renewcommand\twocolumn[1][]{%
    \oldtwocolumn[{#1}{
    
\begin{center}
\vspace{-1.5em}
\includegraphics[width=0.95\textwidth]{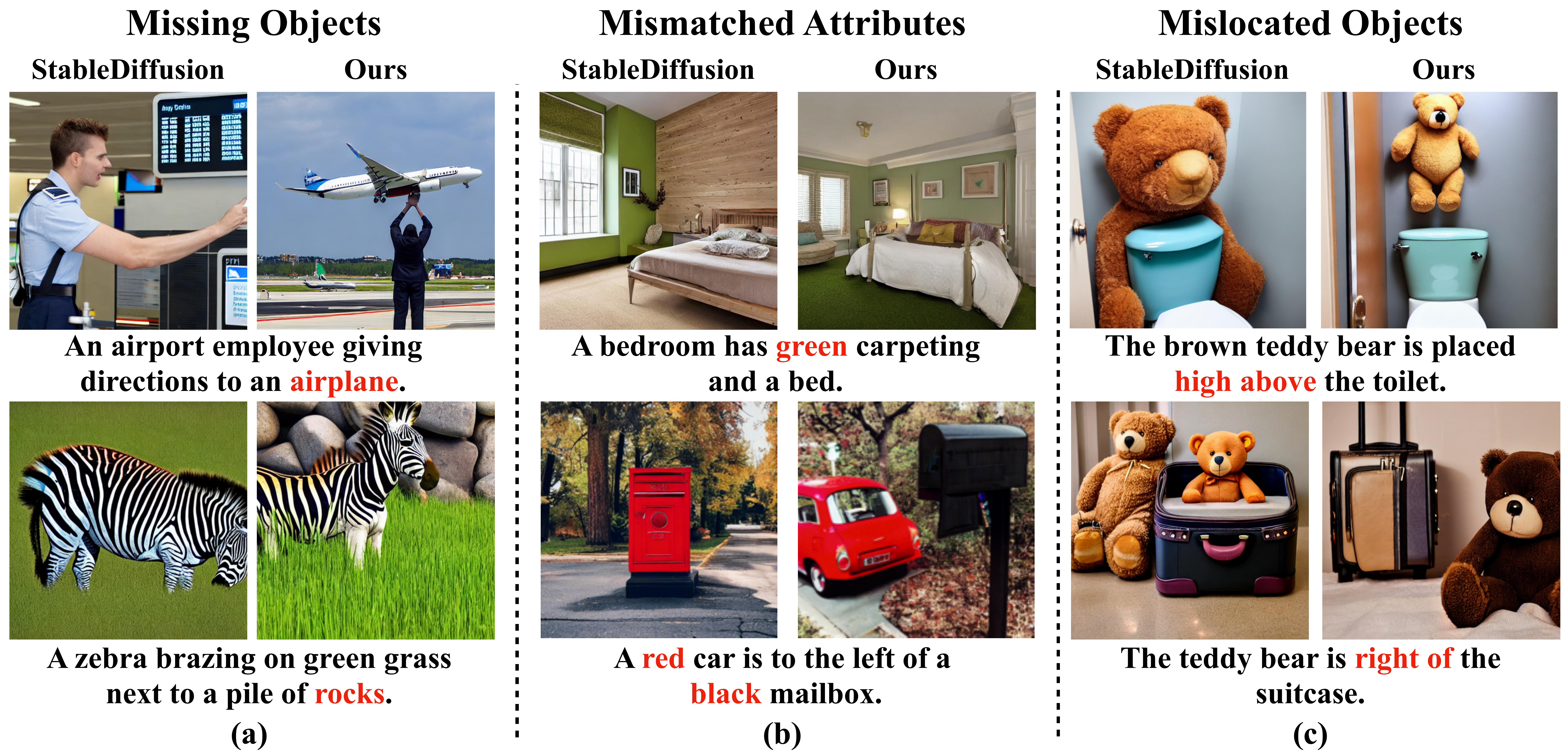}
    \captionof{figure}{
    \textbf{Example generation by stable-diffusion-v1-4 and our method.} Stable diffusion model makes three types of errors that include \textit{missing objects}, \textit{mismatched attributes}, and \textit{mislocated objects}. \textcolor{red}{Errors} are highlighted in red.
    }
    \label{fig:teaser}
\end{center}
    }]
}

\maketitle
\ificcvfinal\thispagestyle{empty}\fi

\begin{abstract}
Diffusion-based models have achieved state-of-the-art performance on text-to-image synthesis tasks.
However, one critical limitation of these models is the low fidelity of generated images with respect to the text description, such as missing objects, mismatched attributes, and mislocated objects.
One key reason for such inconsistencies is the inaccurate cross-attention to text in both the spatial dimension, which controls at what pixel region an object should appear, and the temporal dimension, which controls how different levels of details are added through the denoising steps.
In this paper, we propose a new text-to-image algorithm that adds explicit control over spatial-temporal cross-attention in diffusion models.
We first utilize a layout predictor to predict the pixel regions for objects mentioned in the text. We then impose spatial attention control by combining the attention over the entire text description and that over the local description of the particular object in the corresponding pixel region of that object.
The temporal attention control is further added by allowing the combination weights to change at each denoising step, and the combination weights are optimized to ensure high fidelity between the image and the text.
Experiments show that our method generates images with higher fidelity compared to diffusion-model-based baselines without fine-tuning the diffusion model. Our code is publicly available.\blfootnote{\hspace{-0.5em}$^{*}$Equal contribution.}\footnote{\url{https://github.com/UCSB-NLP-Chang/Diffusion-SpaceTime-Attn}}
\end{abstract}

\section{Introduction}

Diffusion models \cite{ho2020denoising, karras-2022-elucidate, sohl2015deep, song2020denoising, song2020score} have recently revolutionized the field of image synthesis. Compared with previous generative models such as generative adversarial networks \cite{arjovsky2017-wasserstein, brock2018-large, NIPS2014_gan, karras-progressive-gan} and variational autoencoders \cite{kingma2013auto, NEURIPS2019_vq-vae, pmlr-v32-rezende14}, diffusion models have demonstrated superior performance in generating images with higher quality, more diversity, and better control over generated contents. Particularly, text-to-image diffusion models \cite{balaji2022ediffi, nichol2021-glide, ramesh2022hierarchical, rombach2022high, saharia2022photorealistic} allow generating images conditioned on a text description, which enables generation of creative images due to the expressiveness of natural language.

However, recent studies \cite{feng2022training, marcus-preliminary} have revealed that one critical limitation of existing diffusion-model-based text-to-image algorithms is the \emph{low fidelity} with respect to the text descriptions -- the content of the generated image is sometimes at odds with the text description, especially when the description is complex. Specifically, typical errors made by stable diffusion models fall into three categories: \textit{missing objects}, \textit{mismatched attributes}, and \textit{mislocated objects}. 
For example, in Fig.~\ref{fig:teaser}(a), stable diffusion model ignores the airplane even though it is mentioned in text; in Fig.~\ref{fig:teaser}(b), the model confuses ``red car'' and ``black mailbox'' and generates a red mailbox; in Fig.~\ref{fig:teaser}(c), the model locates teddy bear behind the toilet, despite the description ``teddy bear is placed high above the toilet.''

Such infidelity problems suggest that the cross-attention map on the text description may not be accurate. In particular, if we view the generation process of a diffusion model as a sequence of denoising steps, then the cross-attention on text descriptions can be considered as a function of both \emph{spatial} (pixels) and \emph{temporal} (denoising steps) information. Therefore, the inaccuracies of the cross-attention can result from the loose control over both the spatial and temporal dimensions. On one hand, spatial attention controls at what pixels the model should attend to each object and the corresponding attributes mentioned in the text. If the spatial attention is incorrect, the resulting images will have incorrect object locations or miss-associated attributes. On the other hand, temporal attention controls when the models should attend to different levels of details in the text. As previous works have revealed, diffusion models tend to focus on generating object outlines at earlier denoising steps and on details at later \cite{wu2022uncovering}. Thus loose control over the temporal aspect of attention can easily lead to overlooking certain levels of the object details. In short, to improve the fidelity of text-to-image synthesis, one would need to explicitly control both spatial and temporal attention to follow an accurate and optimal distribution.

In this paper, we propose a new text-to-image algorithm based on a pre-trained conditional
diffusion model with explicit control over the spatial-temporal cross-attention map on text. The proposed algorithm introduces a layout predictor and a spatial-temporal attention optimizer. The layout predictor takes the text description as input and generates a spatial layout for each object mentioned in the text. Alternatively, the layout can also be provided by the user. Then the spatial-temporal attention optimizer imposes direct control over the spatial and temporal aspects of the attention according to the spatial layout. In particular, for the spatial aspect, we parameterize the attention map such that the attention outputs in the designated pixel region for an object are a weighted combination of attention over the entire text description and that over the local description that specifically describes the corresponding object. In this way, we manage to emphasize the attention over the object descriptions. For the temporal aspect, we allow the combination weights to change across time and optimized according to a CLIP objective that measures the agreement between the generated images and the text description. 
In this way, we allow the attention to focus more on the entire description at the early stage and gradually shift to the detailed local descriptions as the denoising process proceeds.
The entire pipeline resembles a typical painting process of a human, where each object's position is determined beforehand and the focus gradually shifts from global information to the local details of each object.

We conduct extensive experiments on datasets that contain real and template-based captions \cite{mscoco-caption, liu2022visual} and our newly created synthetic dataset that contains complex text descriptions. Results show that our method generates images that better align with descriptions compared to other stable diffusion-based baselines. As shown in Fig.~\ref{fig:teaser}, our method effectively resolves the above-mentioned three errors. Particularly, controlling spatial attention locates objects at the desired position, and controlling temporal attention promotes the occurrence of objects with associated attributes. Our findings shed light on fine-grained control of diffusion models in text-to-image generation tasks.

\begin{figure*}
    \centering
    \includegraphics[width=0.85\textwidth]{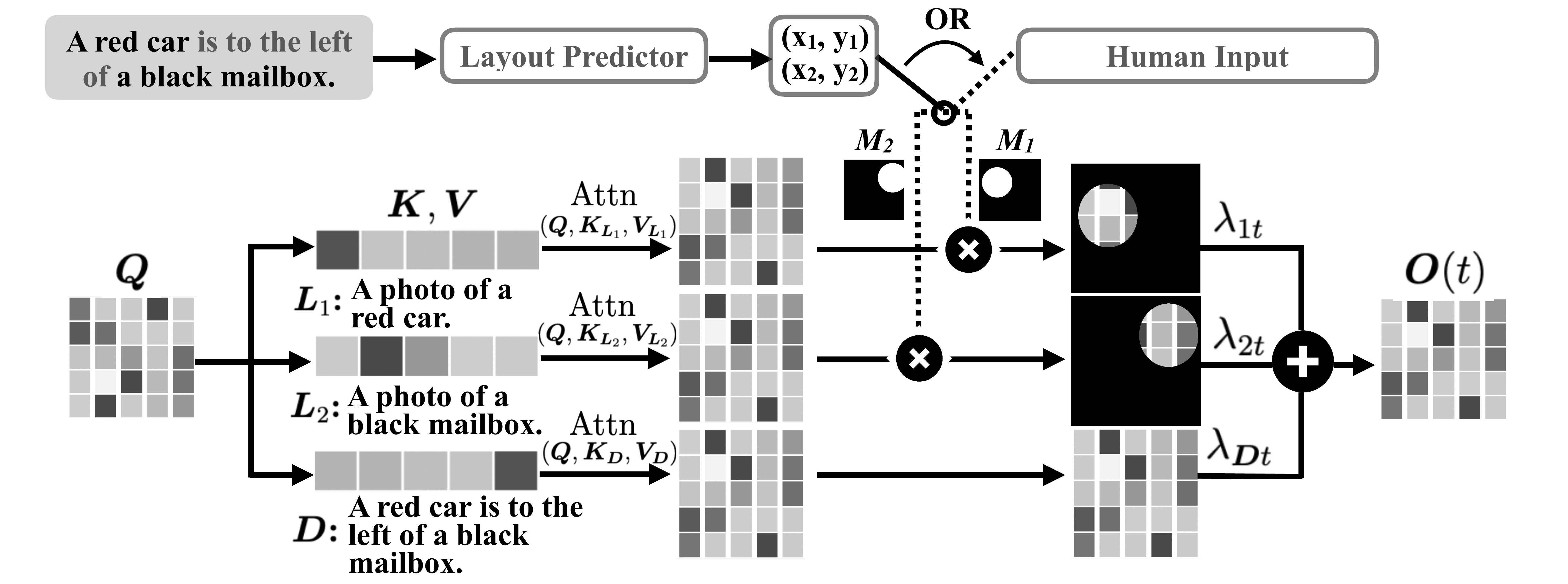}
    \caption{\textbf{Overview of our text-to-image generation pipeline at one denoising step.} Given input text \e{\bm D}, we first parse it and extract all objects mentioned, constructing local descriptions \e{\bm L_i}. Next, the layout predictor predicts the pixel region for each object in the text. The diffusion model attends to the global description \e{\bm D} and additionally attends to the local description \e{\bm L_i} in object \e{i}'s region. The final attention output is a weighted combination of attention to both global and local descriptions, where the combination weights sum up to 1 for each pixel and are optimized for each denoising step to achieve a high fidelity with \e{\bm D}.
    }
    \label{fig:method}
\end{figure*}

\section{Related Work}
\noindent
\textbf{Diffusion Models} \quad Diffusion models are a class of generative models that have demonstrated state-of-the-art performance on image synthesis tasks \cite{dhariwal2021diffusion, ho2020denoising, karras-2022-elucidate, sohl2015deep, song2020denoising, song2020score}.
These models synthesize images by sampling a noisy image from the standard Gaussian distribution and iteratively denoising it back to a clean image.
Their impressive performance has advanced research in multiple computer vision areas including inpainting \cite{lugmayr2022repaint,saharia2022palette,xie2022smartbrush}, image editing \cite{couairon2022diffedit,hertz2022prompt,meng2021sdedit, wu2022uncovering}, super-resolution \cite{ho2022cascaded,saharia2022image}, video synthesis \cite{ho2022imagen,ho2022video,yu2023video}, 
and applications beyond computer vision \cite{kong2020diffwave,li2022diffusion,xu-geodiff}.
Among these, text-to-image diffusion models have gained significant attention \cite{balaji2022ediffi,ramesh2022hierarchical,ramesh2021zero,saharia2022photorealistic,zhou2021lafite}. Taking text descriptions as inputs, these models generate high quality images that are semantically aligned with the input text, which have led to many creative and artistic applications.

\vspace*{0.05in}
\noindent
\textbf{Enhancing the Controllability of Text-to-Image Diffusion Models} \quad
While diffusion-based text-to-image models have shown promising results, recent works have highlighted cases where models fail to generate high-fidelity images with respect to the input text \cite{feng2022training, marcus-preliminary}. To this end, prior works have explored various ways to enhance the controllability of text-to-image diffusion models. One line of work enhances the controllability by improving diffusion models' ability to understand natural language, which includes using a more powerful text encoder that is separately trained on language modeling tasks \cite{balaji2022ediffi,saharia2022photorealistic}, incorporating linguistic structures in the text to guide the cross-attention between image and text \cite{feng2022training}, and decomposing a complex text description into multiple components that are easy to generate \cite{liu2022compositional}.
Another line of work conditions diffusion models' generation on auxiliary inputs such as object layout \cite{balaji2022ediffi, fan2022frido, li2023gligen, yang2022modeling} and silhouette \cite{huang2023region,park2022shape,singh2022high, zeng2022scenecomposer, zhang2023adding}. By modifying diffusion models' attention operation according to these auxiliary information or directly fine-tuning diffusion models to take these auxiliary inputs, they are able to control the location and shape of the objects in the image.
Finally, some work adds temporal aspect control on diffusion models by modifying the input text condition at each denoising step, which allows them to disentangle a desired attribute from other contents \cite{wu2022uncovering}.
Different from prior works, our method imposes both spatial and temporal control in cross-attention layer. Moreover, our method does not require auxiliary inputs and does not fine-tune the diffusion model.

\section{Methodology}

\subsection{Problem Formulation}
\label{sec:formulation}

    We focus on the standard text-to-image problem. Given a text description, denoted as \e{\bm D}, our goal is to generate an image that is consistent with \e{\bm D}. For a concrete exposition, we will use an example description ``\emph{A red car is to the left of a black mailbox}'' in the following. Our work aims to improve the fidelity of the generated image to text, which includes three requirements:
\begin{itemize}[leftmargin=10pt,itemsep=1pt]
    \item \textbf{Object Fidelity:} The generated images should include all the objects mentioned in \e{\bm D}. In our example, the generated image should contain a car and a mailbox.
    \item \textbf{Attribute Fidelity:} The attributes of each object in the image should match those in \e{\bm D}. In our example, the car should be red and the mailbox should be black.
    \item \textbf{Spatial Fidelity:} The relative spatial positions of the object should match the description in \e{\bm D}. In our example, the car should be on the left and the mailbox on the right.
\end{itemize}
We will tackle these problems using a pre-trained, fixed stable diffusion model.

\subsection{Method Overview}
\label{sec:method-overview}

To achieve high-fidelity text-to-image generation, we propose an algorithm consisting of the following four steps.

\noindent
\textbf{Step 1: Object Identification} \quad We extract all the objects mentioned in \e{\bm D}, denoted as \e{O_{1:N}}, by eliciting the noun phrases using spaCy.\footnote{\url{https://spacy.io/}.} \emph{N} is the total number of objects. In our sample, \e{O_1=}\emph{``a red car''} and \e{O_2=}\emph{``a black mailbox''}.

\noindent
\textbf{Step 2: Layout Prediction} \quad For each object \e{O_i}, we use a \emph{layout predictor} to predict its pixel region, \e{\mathcal{R}_i}, which is a set of pixels roughly specifying where the object should be. The layout can also be provided by human users.

\noindent
\textbf{Step 3: Local Description Generation}  \quad For each object \e{O_i}, we generate a local text description, \e{\bm L_i}, containing only that object information using a simple template. In our example, there are two local descriptions. \e{\bm L_1=}\emph{``A photo of a red car''} and \e{\bm L_2=}\emph{``A photo of a black mailbox''}.

\noindent
\textbf{Step 4: Attention Optimization} \quad During the generation process, we guide the diffusion model to combine attention to both the global description \e{\bm D} and the local descriptions \e{\bm L_{1:N}} according to the object layout. The attention combination weights are optimized for each input description.

The following subsections will provide further details about steps 2 and 4.

\subsection{Layout Predictor}
\label{sec:layout-predictor}

Our layout predictor is adapted from \cite{Yang_2021_CVPR_layout}, which aims to predict the center coordinate \e{\bm C_i = [X_i, Y_i]} for each object \e{O_i}. Specifically, the layout predictor is a transformer that takes the text description \e{\bm D} as the input. At the output position where the input mentions the object \e{O_i}, we let the transformer output a set of Gaussian mixture model (GMM) parameters to fit the object's center coordinate \e{\bm C_i}. Formally, denote \e{\bm f_i (\bm D; \bm \theta)} as the output of the layout predictor at the location of the mentioning object \e{O_i}. Then
\begin{equation}
    \small
    \bm f_i (\bm D; \bm \theta) = \bigcup_{k=1}^K \{\bm \mu_{ik}, \bm \Sigma_{ik}, w_{ik}\},
\end{equation}
where \e{\bm \mu_{ik}}, \e{\bm \Sigma_{ik}}, and \e{w_{ik}} denote the mean, covariance matrix and the prior probability of mixture \e{k}; \e{\bm \theta} denotes the network parameters.

To improve the fidelity of the object positions in the predicted layout, we introduce a hybrid training objective including an \emph{absolute position objective} and a \emph{relative position objective}.

\vspace*{0.1in}
\noindent
\textbf{Absolute Position Objective} \quad The absolute position objective provides direct supervision of the exact position of each object. Formally, we assume access to an image captioning dataset \e{\mathcal{D}_{\textrm{real}}} with description \e{\bm D} as well as the extracted labels of all the objects \e{\{O_i\}} and their center coordinates \e{\{\bm C_i\}}. Then the training objective is to minimize the negative log-likelihood of the ground-truth coordinates under the predicted GMM distribution, \emph{i.e.}
\begin{equation}
\small
    \mathcal{L}_{\textrm{abs}}(\bm \theta) = \mathbb{E}_{\bm D, \{\bm Ci\} \sim \mathcal{D}_{\textrm{real}}} \Big[\sum_{i=1}^N\ell_{\textrm{nll}}( \bm C_i; \bm f_i(\bm D; \bm \theta))\Big],
\end{equation}
where \e{\ell_{\textrm{nll}}(\bm C_i; \bm f_i)} denotes the negative log-likelihood of ground truth \e{\bm C_i} under the GMM specified by \e{\bm f_i}.

\vspace*{0.1in}
\noindent
\textbf{Relative Position Objective} \quad In many cases, the text description only mentions the relative positions of the objects, and thus it is more important to ensure the relative position of the predicted position is correct than the absolute position. To further enforce the fidelity of the relative positions, we introduce the following relative position objective.

To start with, we construct a synthetic dataset, \e{\mathcal{D_{\textrm{syn}}}}, which consists of text description \e{\bm D} with explicit descriptions of relative spatial relations. We first randomly select \e{N} objects with attribute modifiers. We then select \e{M} pairs of objects to specify their relative spatial relation. For object pair \e{(O_i, O_j)}, their spatial relation, \e{R_{ij}}, is randomly drawn from ``left of'', ``right of'', ``above'' and ``below''. Finally, we prompt GPT-3 \cite{gpt3} to generate the text description \e{\bm D} that mentions all the objects and their relative positions. Our familiar ``A red car is to the left of a black mailbox'' is one such example. Further details are provided in Appendix~\ref{Append:datasets}.

Next, we introduce a loss to penalize violating the relative position. If object \e{i} is to the left of object \e{j}, we enforce that the rightmost mixture mean of object \e{i} is to the left of the leftmost mixture mean of object \e{j}. Formally,
\begin{equation}
    \small
    \ell_{\textrm{rel}}(R_{ij}=\textrm{``left''}; \bm f_i, \bm f_j) = \max\{\max_k \bm \mu_{ik}(0) - \min_k \bm \mu_{jk}(0), -\delta \},
\end{equation}
where \e{\bm \mu_{ik}(0)} denotes the zeroth element (the \e{x}-coordinate) of \e{\bm \mu_{ik}}, and \e{\delta} is a pre-specified margin. The \e{\ell_{\textrm{rel}}} of the other three types of relations are defined similarly. The relative position loss is thus the aggregation of \e{\ell_{\textrm{rel}}} across the synthetic dataset:
\begin{equation}
    \small
    \mathcal{L}_{\textrm{rel}}(\bm \theta) = \mathbb{E}_{\bm D, \{R_{ij}\} \sim \mathcal{D}_{\textrm{syn}}} \Big[\sum_{i,j:R_{ij}\neq \emptyset} \ell_{\textrm{rel}}(R_{ij}; \bm f_i, \bm f_j) \Big],
\end{equation}
where \e{\bm f_i} is short for \e{\bm f_i(\bm D; \bm \theta)}.

\vspace*{0.1in}
\noindent
\textbf{Training and Inference} \quad To sum up, the final training objective is the combination of both:
\begin{equation}
    \small
    \mathcal{L}_{\textrm{layout}}(\bm \theta) = \mathcal{L}_{\textrm{abs}}(\bm \theta) + \xi \mathcal{L}_{\textrm{rel}}(\bm \theta),
\end{equation}
where \e{\xi} is a hyperparameter. During inference time, we randomly draw the center coordinate \e{\bm C_i} from the predicted GMM. The pixel region for object \e{O_i}, \e{\mathcal{R}_i}, is defined as a circular region centered at the drawn \e{\bm C_i} with a \emph{fixed radius} \e{r}. As would be shown in Appendix~\ref{Append:layout-effect}, \e{\mathcal{R}_i} only roughly regulates the position of the generated objects, and the actual size of the object can go beyond or below the size of \e{\mathcal{R}_i}. Thus a fixed \e{r} is sufficient for this purpose.

\subsection{Spatial-Temporal Attention Optimization}
\label{sec:spatial-temporal}

Recall that \e{\bm D} is the global description and \e{\{\bm L_i\}} are local descriptions for each object. Our goal is to guide the diffusion model to attend to not only \e{\bm D}, but also \e{\bm L_i} in object \e{i}'s region, \e{\mathcal{R}_i}, so that the model is more strongly prompted to generate the specific object as specified in the layout.

\vspace*{0.1in}
\noindent
\textbf{Spatial-Temporal Attention} \quad Recall that the standard cross-attention \cite{vaswani-attention} is defined as
\begin{equation}
    \small
    \textrm{Attention}(\bm Q, \bm K, \bm V) = \textrm{softmax}\big(\bm Q \bm K^T / \sqrt{d}\big) \bm V,
\end{equation}
where \e{\bm Q \in \mathbb{R}^{h\times w, d}} is the queries vectors for each \emph{pixel}, and \e{\bm K, \bm V \in \mathbb{R}^{l, d}} are key and value vectors for each \emph{text description token}; \e{h} and \e{w} represent the image height and width, \e{l} represents the text description length; \e{d} represents the dimension of each attention vector.

Since we have multiple text descriptions, we define \e{\bm K_{\bm D}, \bm V_{\bm D}} as the key and value vectors for the global description \e{\bm D}, and \e{\bm K_{\bm L_i}, \bm V_{\bm L_i}} for the local description \e{\bm L_i}. Further, we introduce a set of binary mask matrices \e{\{\bm M_i\}} to indicate the region for each object, \emph{i.e.}
\begin{equation}
\small
    \bm M_i(x,y) = 1, \mbox{ if } (x, y) \in \mathcal{R}_i, \quad \mbox{and } 0, \mbox{ otherwise}.
\end{equation}
Then the output of the attention layer of the denoising network at time \e{t} is defined as
\begin{equation}
\vspace*{-0.05in}
\small
\begin{aligned}
    \bm O(t) &= \sum_{i=1}^N \lambda_{it} \bm M_i \odot \textrm{Attention}(\bm Q, \bm K_{\bm L_i}, \bm V_{\bm L_i}) \\
    &+ \Big(1 - \sum_{i=1}^N \lambda_{it} \bm M_i\Big) \odot \textrm{Attention}(\bm Q, \bm K_{\bm D}, \bm V_{\bm D}),
\end{aligned}
\label{eq:optimizer}
\end{equation}
where \e{\lambda_{it}} are the attention combination weights, and \e{\odot} denotes element-wise multiplication. Note that the combination weights are functions of time \emph{t}, which is motivated by the observation that different denoising steps control the generation of different levels of details. By introducing this time dependency, we allow the diffusion model to focus more on the global description at earlier time steps and shift to local descriptions later.

\vspace*{0.1in}
\noindent
\textbf{Optimization Objective} \quad All the attention combination weights, \e{\bm \lambda = \{\lambda_{it}\}}, are determined by maximizing the consistency between the generated image and the text description as measured by the CLIP similarity \cite{radford-clip-2021}. We introduce two CLIP similarities, a global CLIP similarity that compares the entire image and the global description, and a set of local CLIP similarities that compare the images at each object region and the corresponding local descriptions. Formally, denote the generated image as \e{\bm I(\bm \lambda)}, and its local patch at each object's region as \e{\bm I_{O_i}(\bm \lambda)}.\footnote{To standardize the image patch sizes, $\bm I_{O_i}$ is obtained by cropping the original image with a minimum square that encompasses $\mathcal{R}_i$ and resizing it to 224$\times$224.} Note that we adopt the \emph{deterministic} PLMS denoising process \cite{songpseudoinverse} (\emph{i.e.}, with \e{\sigma = 0}), so both \e{\bm I} and \e{\bm I_{O_i}} are deterministic functions of \e{\bm \lambda}. Then the loss for optimizing \e{\bm \lambda} is given by
\begin{equation}
    \small
    \mathcal{L}_{\textrm{attend}}(\bm \lambda) = -\textrm{CLIP}(\bm I(\bm \lambda), \bm D) - \gamma \sum_{i=1}^N \textrm{CLIP}(\bm I_{O_i}(\bm \lambda), \bm L_i),
\end{equation}
where \e{\textrm{CLIP}(\cdot)} denotes the CLIP similarity, and \e{\gamma} is a hyperparameter.

\section{Experiments}\label{exp}
We conduct experiments to evaluate our method's performance and generalizability. We also perform ablation study on important design choices of our method.

\vspace*{0.05in}
\noindent\textbf{Implementing Details:}
We adopt RoBERTa-base \cite{liu2019roberta} as the base model for layout predictor and use 5 mixtures in GMM. We use stable-diffusion-v1-4 \cite{rombach2022high} pre-trained on the laion dataset \cite{schuhmann2021laion} and freeze it throughout all experiments. All generated images are in the size of $512 \times 512$. We use PLMS sampler \cite{songpseudoinverse} to synthesize images with 50 denoising steps, and we use Adam \cite{kingma2014adam} to optimize the layout predictor and attention combination weights. More details on hyperparameters and optimization are in Appendix \ref{Append:implement-detail}.

\subsection{Evaluation on Fidelity of Generated Images}
\label{sec:main-results}

\begin{table*}
    \centering
    \resizebox{1.0\linewidth}{!}{%
\begin{tabular}{lcccccccccc}
\toprule
\multicolumn{1}{c}{\multirow{3}{*}{\textbf{}}} & \multicolumn{3}{c}{\textbf{MS-COCO}} & \multicolumn{3}{c}{\textbf{VSR}} & \multicolumn{4}{c}{\textbf{GPT-synthetic}}\\

\cmidrule(lr){2-4}
\cmidrule(lr){5-7}
\cmidrule(lr){8-11}

  & Object $(100)\uparrow$ & Attribute $(100)\uparrow$ & Overall
  & Object $(100)\uparrow$ & Spatial $(100)\uparrow$ & Overall
  & Object $(100)\uparrow$ & Attribute $(100)\uparrow$ & Spatial $(100)\uparrow$ & Overall \\

\midrule

\textsc{Vanilla-SD}             & 66 & 62 &  $42\%$ & 64 &  66 & $48\%$ & 60 & 58 & 60 & $38\%$ \\
\textsc{Composable-Diffusion} & 53 & 48 &  $30\%$ & 63 &  60 & $18\%$ & 63 & 60 & 47 & $28\%$ \\
\textsc{Structure-Diffusion}  & 60 & 63 &  $48\%$ & 54 &  64 & $32\%$ & 58 & 54 & 58 & $32\%$ \\
\textsc{Paint-with-Words}     & \textbf{75} & \textbf{74} & $52\%$ & 54  & 59 & $30\%$ & 65 & 52 & 57 & $32\%$ \\
\rowcolor{gray!20}
Ours                          & \textbf{75} & \textbf{74} &  -- & \textbf{68}  & \textbf{77} & -- & \textbf{84} & \textbf{83} & \textbf{86} & -- \\

\bottomrule

\end{tabular}
}
\caption{\textbf{Subjective evaluation of our method and baselines.} \textbf{Best numbers} are in bold. Spatial relation is not available on MS-COCO because very few of its captions contain spatial relations. Attribute is not available on VSR as its captions do not consider attribute. Object, Attribute, and Spatial show the total score of 50 evaluations, where a model with the highest fidelity would achieve a score of 100. Overall denotes the percentage of generation of each method that is rated better than our method.
}
\vspace*{-0.1in}
\label{tab:subjective_eval}
\end{table*}

\begin{table}
    \centering
    \resizebox{1.0\linewidth}{!}{%

\begin{tabular}{lcccccccccccc}
\toprule
\multicolumn{1}{c}{\multirow{3}{*}{\textbf{}}} & \multicolumn{2}{c}{\textbf{MS-COCO}} & \multicolumn{2}{c}{\textbf{VSR}} & \multicolumn{2}{c}{\textbf{GPT-synthetic}}\\

\cmidrule(lr){2-3}
\cmidrule(lr){4-5}
\cmidrule(lr){6-7}

  & Object   & SPRel  
  & Object   & SPRel  
  & Object   & SPRel   \\

  & Recall   & Precision  
  & Recall   & Precision  
  & Recall   & Precision  \\

\midrule
\textsc{Vanilla-SD} & 58.0\% & -- & 62.1\% & 56.0\% & 42.4\% & 56.8\% \\
\textsc{Composable-Diffusion} & 51.8\% & -- & 60.8\% & 72.2\% & 34.4\% & 52.6\%\\
\textsc{Structure-Diffusion} & 61.7\% & -- & 62.3\% & 58.3\% & 43.0\% & 59.3\%\\
\textsc{Paint-with-Words} & 57.3\% & -- & 63.8\% & 	47.1\% & 45.9\% &  53.5\% \\
\rowcolor{gray!20}
Ours & \textbf{69.6\%} & -- & \textbf{65.1\%} & \textbf{75.0\%} & \textbf{47.2\%} & \textbf{66.7\%}\\

\bottomrule

\end{tabular}
}
\caption{\textbf{Automatic evaluation of our method and baselines.} SPRel Precision: Spatial Relation Precision. \textbf{Best numbers} are in bold. SPRel precision is not available on MS-COCO since most captions do not have an explicit spatial relation.
}
\vspace*{-0.1in}
\label{tab:main_results}
\end{table}

We first evaluate our method on object, attribute, and spatial fidelities as introduced in Sec.~\ref{sec:formulation} using both objective and subjective metrics.

\vspace*{0.05in}
\noindent
\textbf{Baselines:} We identify four baseline methods on generating images from complex text descriptions. (1) \textsc{Vanilla-SD} \cite{rombach2022high} is the pre-trained text-to-image stable diffusion model that directly generates images conditioned on the text description.
(2) \textsc{Composable-Diffusion} \cite{liu2022compositional} is a diffusion-based compositional generation method. To generate images from a text description, the text is first decomposed into the conjunction of multiple components (\emph{e.g.,} for the example in Sec.~\ref{sec:formulation}, the components are ``A red car is to the left of a black mailbox.'' AND ``A red car'' AND ``a black mailbox''). Each component is separately modeled by the diffusion model and then composed to generate an image by merging the outputs of each denoising step.
(3) \textsc{Structure-Diffusion} \cite{feng2022training} improves \textsc{Vanilla-SD} on generating images with better object and attribute fidelities. They do so by first extracting noun phrases at different levels of the parsing tree. The model then separately attends to each noun phrase and combines the attention outputs by taking their arithmetic mean. Different from our method, they perform cross-attention on the whole image instead of the region specific to an object.
(4) \textsc{Paint-with-Words} \cite{balaji2022ediffi} assumes that users provide the pixel region of each object to be generated. Pixels in the region then increase their attention weights to the text that describes the corresponding object, and the amount of increase is determined by heuristic rules. For a fair comparison, we use the pixel region predicted by our layout predictor, and we report the performance with the ground truth region in Appendix~\ref{Append:ablation-layout}.

\vspace*{0.05in}
\noindent
\textbf{Datasets and Metrics:} We conduct experiments on three datasets. (1) \textbf{MS-COCO} \cite{lin2014microsoft} contains photos taken by photographers and manually annotates the caption for each photo. We use the caption as the input text description. (2) \textbf{VSR} \cite{liu2022visual} is proposed for probing spatial understanding of vision-language models. Constructed from a subset of MS-COCO, it uses templates to generate captions that describe spatial relations in the image (\emph{e.g.,} ``The horse is to the left of the person.''). (3) \textbf{GPT-synthetic} is the synthetic dataset introduced in Sec.~\ref{sec:layout-predictor}. We manually check the test set to filter out sentences that do not conform to the specified spatial relation. Compared with VSR, this dataset has more diverse and complex text descriptions. The final dataset contains 500 descriptions, and we downsample MS-COCO and VSR to have the same size. Statistics including the number of objects and spatial relations are shown in Appendix \ref{Append:datasets}.

For automatic evaluations, we consider two metrics. (1) \textbf{Object Recall} measures the percentage of successfully synthesized objects over objects mentioned in the text. We use DETR \cite{carion-detr} to detect objects in generated images. To calculate recall, we divide the number of detected objects in the text by the total number of objects in the text that also belong to one of the MS-COCO categories. It measures the \textbf{object fidelity} of generated images.
(2) \textbf{Spatial Relation Precision} (SPRel Precision) measures the percentage of the correct spatial relations among all the relations whose corresponding objects are successfully synthesized.
This metric measures the \textbf{spatial fidelity} of generated images.
We consider the relation of left, right, above, and below because their correctness can be evaluated by comparing the bounding box centers. Qualitative examples of spatial relations beyond these four are shown in
Appendix~\ref{Append:qualitative-examples}.
We also report the CLIP similarities between generated images and the input description in Appendix~\ref{Append:CLIP}. 
We observe that different methods have very close CLIP similarities, though our method still achieves competitive results.
\begin{figure*}
\centering
\vspace{-0.04in}
\includegraphics[width=0.93\textwidth]{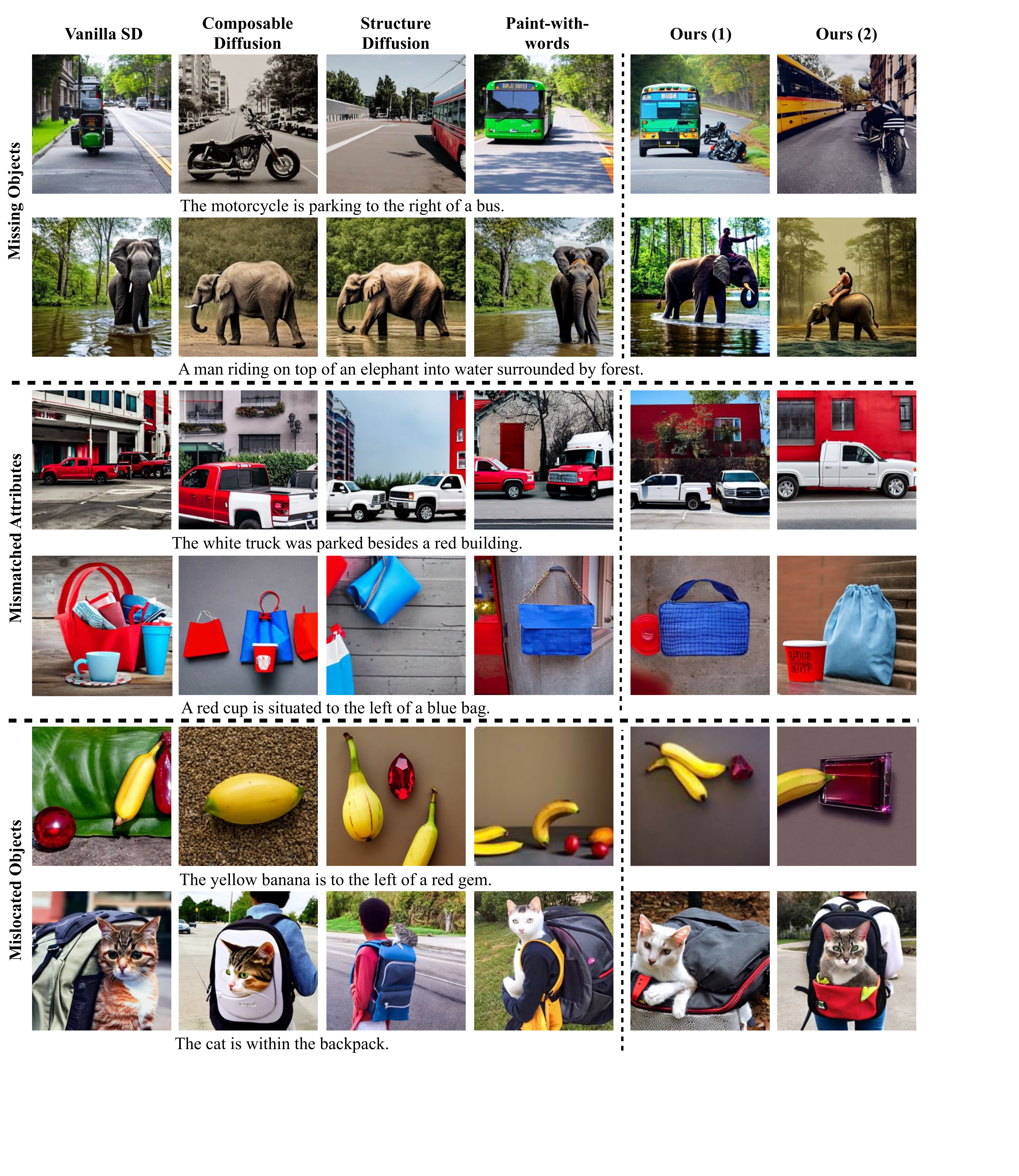}
\vspace{-0.04in}
\caption{\textbf{Example images generated by our method and baselines.} Typical errors of baselines include missing objects, mismatched attributes, and mislocated objects. Ours (1)/(2) show the results with two different random seeds.}
\vspace*{-0.05in}
\label{fig:qualitative-analysis}
\end{figure*}

\vspace*{0.05in}
\noindent
\textbf{Results:} The results are shown in Table~\ref{tab:main_results}. As shown in the table, our method outperforms all baselines on both object recall and SPRel precision. Specifically, compared with baselines that do not have spatial attention control (all but \textsc{Paint-with-Words}), 
our method is significantly better on SPRel precision, which illustrates the effectiveness of spatial control and indicates that it is difficult for the stable diffusion model to generate correct spatial relations without guidance on the layout. Our method moves the burden of generating objects at the correct location from diffusion models to a separate layout predictor, allowing the whole pipeline to achieve better spatial fidelity.
On the other hand, compared with \textsc{Paint-with-Words} that does not impose the temporal attention control, our method achieves better object recall, showing that our optimized combination weights across the temporal dimension strike a better balance between local and global descriptions.

Fig.~\ref{fig:qualitative-analysis} demonstrates examples of text descriptions and images generated by our method and baselines. We observe that our method resolves three types of errors in baselines.
First, our method alleviates the missing object issue (in the top panel). Baselines tend to focus on one object in the text and ignore other objects (\emph{e.g.,} focusing on the elephant and ignoring the man in the second row), whereas our method generates all objects.
Second, as shown in the middle panel, our method mitigates the mismatched attribute issue.
Particularly, baselines struggle when multiple objects are mentioned in the text, where they mismatch the attribute and object (\emph{e.g.,} red truck in the first row).
Finally, the bottom panel shows examples where our method reduces the number of mislocated objects. Note that our method is effective both on the four relations and other spatial relations. More examples can be found in Appendix \ref{Append:qualitative-examples}.

\vspace*{0.05in}
\noindent
\textbf{Subjective Evaluation:}
To further evaluate the fidelity and quality of the generated images, we perform a subjective evaluation on Amazon Mechanical Turk. Specifically, we randomly sample 25 text descriptions for each dataset in Table \ref{tab:main_results}. Each subject was presented with the text description and corresponding image generated by a single method or a pair of methods and asked the following four questions:
(1) (\textit{Object Fidelity}) Does the image contain all objects mentioned in the text? (2) (\textit{Attribute Fidelity}) Are all synthesized objects consistent with their characteristics described in the text (\emph{e.g.,} color and material)? (3) (\textit{Spatial Fidelity}) Does the image locate all objects at the correct position such that the spatial relations in the text are satisfied (if an object in the relationship is missing, it is considered as an incorrect generation)? and (4) (\textit{Overall}) Which image in the pair has higher fidelity with the text and has better quality? The first three questions are evaluated for each method individually with a score of 0, 1, or 2, where 2 denotes all objects/attributes/relations are correct and 0 denotes none of them is correct. The last question is evaluated on a pair of images, one generated by our method and the other by a baseline.
Each generated image is evaluated by two subjects, so the total score of the first three questions is 100. More details are in Appendix \ref{Append:subjective-eval}.

Table \ref{tab:subjective_eval} shows the results.
Our method achieves significantly better performance in most cases, especially on GPT-synthetic dataset that contains multiple objects and relations in the same description.
The results show that our method is effective at generating images with high fidelity without sacrificing perceptual quality.

\subsection{Additional Analyses}
\label{sec:addition-analysis}
\begin{figure}
    \centering
    \vspace{-0.07in}\includegraphics[width=0.96\columnwidth]{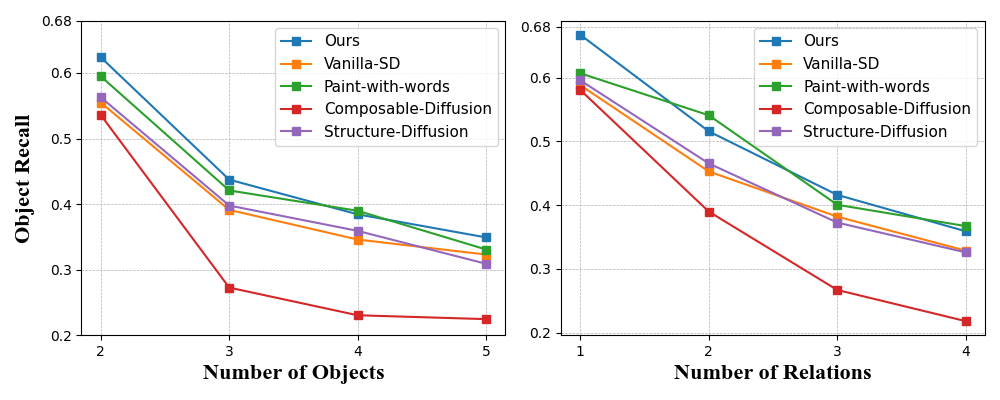}
    \vspace*{-0.02in}
    \caption{Performance when the number of objects and spatial relations in the text increase.}
    \label{fig:multi-object}
\end{figure}

\noindent
\textbf{Performance on Each Text Complexity Level:}
To explore the capability of our method, especially on complex text, we analyze the performance at each complexity level. Specifically, we use the GPT-synthetic dataset where multiple objects and relations can appear in the same description. We use the number of objects \e{N} and number of spatial relations \e{M} introduced in Sec.~\ref{sec:layout-predictor} as proxies
for text complexity and plot the performance of each method as the text becomes more complex.
\emph{i.e.,} we plot the performance for each value of \e{N} regardless of \e{M}, and vise versa for \e{M}.
As shown in Fig.~\ref{fig:multi-object}, the performance of all methods decrease as the text becomes more complex, but our method still outperforms or is on par with others on complex descriptions. Note that \textsc{Paint-with-Words} uses the same layout predictor as ours, which partially explains its strong performance.

\begin{figure}
\centering
\includegraphics[width=0.5\textwidth]{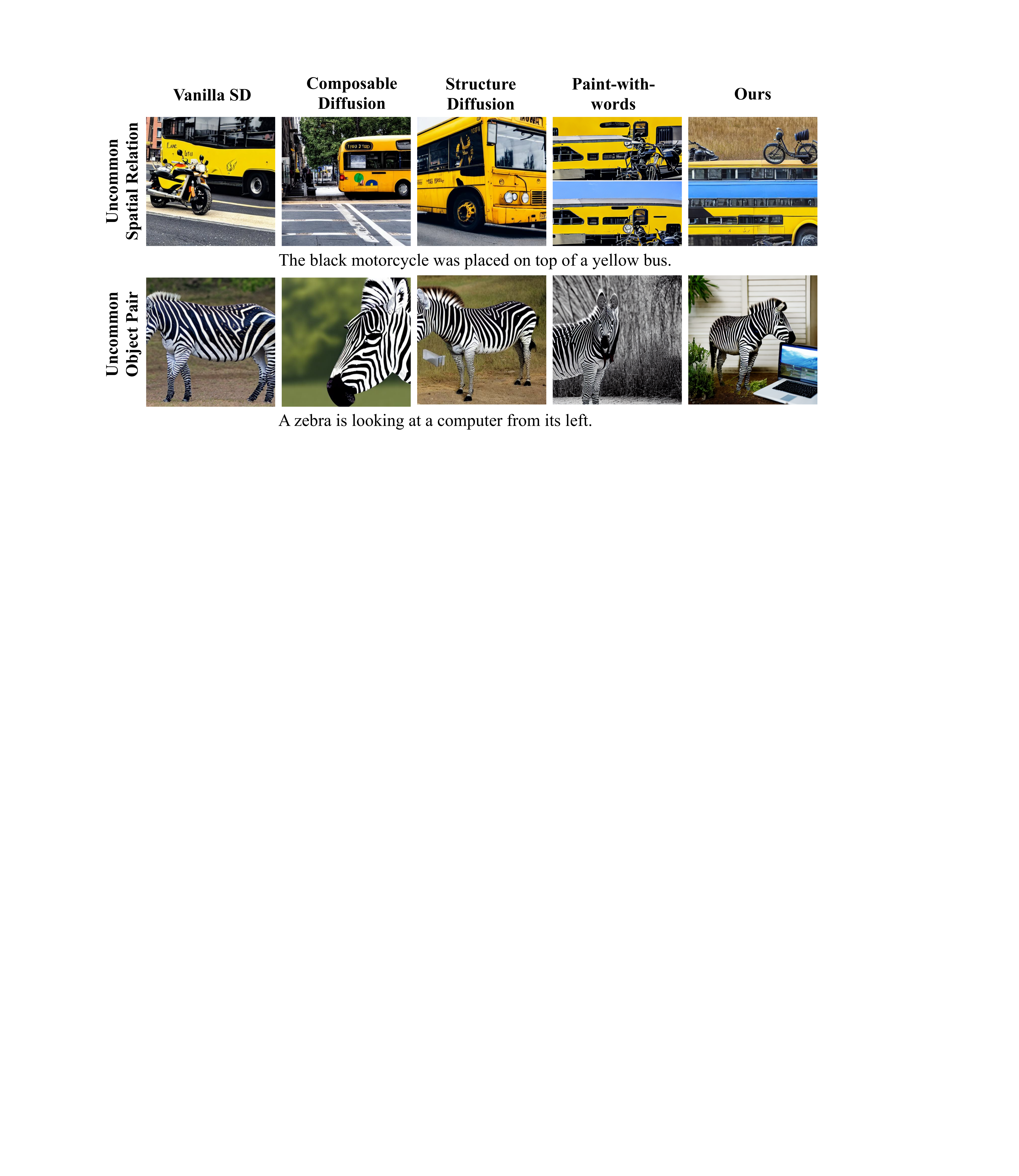}
\caption{Example images generated by our method and baselines on novel object combinations.}
\vspace*{-0.05in}
\label{fig:rare-combination}
\end{figure}

\vspace*{0.05in}
\noindent
\textbf{Generalizability to novel object combinations}
is a critical requirement for text-to-image models. However, since MS-COCO dataset is collected from real photos, most of its captions contain object pairs and object-attribute pairs that are common in daily life, which are also more likely to overlap with stable diffusion's pre-training data. Thus we additionally evaluate our method on another synthetic dataset that is created similarly to Sec.~\ref{sec:layout-predictor}, but contains \textit{uncommon} object pairs, object-attribute pairs, and object spatial relations. Fig.~\ref{fig:rare-combination} shows sample images generated by our method and baselines. It can be observed that our method successfully generates novel object pairs (\emph{e.g.,} ``zebra is looking at computer'').
Performance on this dataset and more generated examples are shown in Appendix~\ref{Append:rare-objects}. Overall, our method is able to generalize to novel object combinations.

\subsection{Ablation Study}
\label{sec:ablation}

\begin{table}
\centering
\resizebox{0.85\linewidth}{!}{%

\begin{tabular}{lcc}
\toprule

  & Object Recall   &  SPRel Precision \\
\midrule
Ours & \textbf{47.2\%} & \textbf{66.7\%} \\
\midrule
No spatial control & 39.6\% & 51.7\%\\
No temporal control & 43.8\% & 61.4\%\\
No optimization & 41.0\% & 55.6\%\\
\bottomrule

\end{tabular}
}
\caption{Ablation performance on GPT-synthetic test set.}
\vspace*{-0.1in}
\label{tab:ablation-attention}
\end{table}

In this section, we will investigate the influence of two important steps in our method, namely, the layout predictor and the spatial-temporal attention. We first study how spatial-temporal attention affects our performance. To do that, we consider three variants of attention.

First, we explore the attention \textbf{without spatial control}. Concretely, the diffusion model still attends to both global description and local descriptions. However, instead of only attending to a local description in the pixel region of the object, the model now attends to all local descriptions in the \textit{whole image}. The combination weights at each denoising step are optimized with only the global CLIP similarity in Sec.~\ref{sec:spatial-temporal}. Table~\ref{tab:ablation-attention} shows that its performance drops drastically, demonstrating the importance of spatial control.

Second, we explore the attention \textbf{without temporal control}, where combination weights remain the same for all denoising steps. Note that the weights can be different for each object and are optimized with \e{\mathcal{L}_{\textrm{attend}}(\bm \lambda)} in Sec.~\ref{sec:spatial-temporal}. The results in Table~\ref{tab:ablation-attention} show a significant degradation under both metrics, indicating the benefits of temporal dependency on attention.

Finally, we explore the attention \textbf{without optimization}. Specifically, we fix the combination weight \e{\lambda_{it} = \frac{1}{N}} for all \e{i} and \e{t}, where \e{N} is the number of objects. This value is also the initialization point of \e{\{ \lambda_{it} \}} in our method. As shown in Table~\ref{tab:ablation-attention}, the performance further drops compared to the no temporal control variant. The results indicate that optimizing the combination weights for a new text description is critical for generating images with high fidelity.

We further study the effects of our layout predictor, which includes the layout predictor trained with only one of the absolute and relative position objectives, the comparison with ground truth and user-provided pixel region, and a different strategy to construct the pixel region for an object. Results of these experiments are presented in Appendix~\ref{Append:ablation-layout}.

\section{Conclusion}
In this work, we study the text-to-image synthesis task based on diffusion models. We find existing methods lack explicit control on cross-attention in diffusion models, which leads to the generation of low-fidelity images. We propose an algorithm that imposes control on cross-attention in spatial and temporal aspects. Experiments show our method outperforms baselines in generating high-fidelity images. Further ablation study verifies the effectiveness of our spatial and temporal attention control. One limitation of our method is the reliance on a time-consuming optimization scheme, which takes around 10 minutes for each text-to-image generation. Future work may consider striking a better balance between performance and efficiency.

\clearpage

{\small
\bibliographystyle{ieee_fullname}
\bibliography{egbib}
}
\newpage
\clearpage
\begin{alphasection}

\begin{table}[b]
\vspace*{-0.05in}
\centering
\resizebox{0.8\linewidth}{!}{%
\begin{tabular}{lcccc}
\toprule
  & 2 objects  & 3 objects & 4 objects & 5 objects \\
\midrule
1 relation & 200 & 50 & 0 & 0 \\
2 relations & 0 & 50 & 50 & 0\\
3 relations & 0 & 0 & 50 & 50\\
4 relations & 0 & 0 & 0 & 50\\

\bottomrule
\end{tabular}
}
\caption{Statistics of GPT-synthetic dataset.}
\label{tab:append-data-stats}
\end{table}

\begin{table}[!h]
    \centering
    \resizebox{1.0\linewidth}{!}{%

\begin{tabular}{lcccccccccccc}
\toprule
\multicolumn{1}{c}{\multirow{3}{*}{\textbf{}}} & \multicolumn{2}{c}{\textbf{MS-COCO}} & \multicolumn{2}{c}{\textbf{VSR}} & \multicolumn{2}{c}{\textbf{GPT-synthetic}}\\

\cmidrule(lr){2-3}
\cmidrule(lr){4-5}
\cmidrule(lr){6-7}

  & Global   & Local  
  & Global   & Local  
  & Global   & Local   \\

  & CLIP   & CLIP  
  & CLIP   & CLIP  
  & CLIP   & CLIP  \\

\midrule
\textsc{Vanilla-SD} & 0.2890 & 0.2357 & \textbf{0.3071} & 0.2412 & \textbf{0.3006}& 0.2243 \\
\textsc{Composable-Diffusion} & 0.2892 & \textbf{0.2397} & 0.2948 & 0.2394 & 0.2886 & 0.2327\\
\textsc{Structure-Diffusion} & 0.2870 & 0.2339 & 0.2972 & 0.2395 & 0.2912 & 0.2396\\
\textsc{Paint-with-Words} & \textbf{0.2902} & 0.2391 & 0.2974 & 0.2394 & 0.2961 & \textbf{0.2418} \\
\rowcolor{gray!20}
Ours & 0.2892 & 0.2375  & 0.3029 & \textbf{0.2415} & 0.2944 & 0.2403\\

\bottomrule

\end{tabular}
}
\caption{CLIP Similarity of our method and baselines.}
\vspace*{-0.1in}
\label{tab:CLIP_similarity}
\end{table}

\begin{table*}[!t]
    \centering
    \resizebox{0.65\linewidth}{!}{%
    \begin{tabular}{lll}
    \toprule
    Module & Attribute  & Value \\
    \midrule
    \multirow{8}{*}{\textbf{Layout Predictor}} & Model checkpoint  & \texttt{Roberta-base}~\cite{liu2019roberta} \\
        & Layers & 12 \\
        & Heads & 12  \\
        & Hidden dimension \e{d} & 768 \\
        & Training batch size & 64 \\
        & Training epoch & 100 \\
        & \multirow{2}{*}{Learning rate} & \e{1e-6 \rightarrow 1e-8} for transformer layer \\
        & & \e{4e-5 \rightarrow 1e-8} for GMM \\
    \midrule
    \multirow{7}{*}{\textbf{Diffusion Model}} & Model checkpoint  & \texttt{stable-diffusion-v1-4}~\cite{rombach2022high} \\
        & Sampling steps & 50\\
        & Sampling variance & 0.0\\
        & Resolution & $512\times 512$\\
        & Latent channels & 4 \\
        & Latent down-sampling factor & 8 \\
        & Conditional guidance scale & 7.5 \\
     \midrule
    \multirow{5}{*}{\textbf{Attention Optimization}} & Checkpoint for CLIP loss &  \texttt{ViT-B/32}~\cite{radford-clip-2021} \\
        & \multirow{1}{*}{\e{\gamma}}  & 5\\
        & Optimizer & Adam~\cite{kingma2014adam} \\
        & Learning rate & 0.05\\
        & \multirow{1}{*}{\e{\lambda_t} initialization} & \e{1/N}, where \e{N} is object numbers\\
    \bottomrule
    \end{tabular}
    }
    \caption{Hyperparameters and model architectures used in this paper.}
    \label{tab:supp-hyperparameters}
    \vspace*{-0.05in}
\end{table*}

\begin{table*}[!t]
    \centering
     \resizebox{0.85\linewidth}{!}{%
    \begin{tabular}{ll}
    \toprule
    \multirow{3}{*}{\textbf{Instruction}} & Given several objects, write a sentence that describes the given objects. Additionally, if location relation \\ & between objects is specified, the sentence needs to contain sufficient information that reveals the relation. \\ & Try to generate sentences as diverse as possible and DO NOT simply state the object locations.  \\
    \midrule
    \multirow{5}{*}{\textbf{Demonstrations}} & Objects: silver car, green motorcycle, blue bus, yellow truck \\
    & Relation: silver car right of blue bus, yellow truck left of blue bus\\
    & Sentence: The blue bus was driving along the road, with a silver car positioned to its right and a yellow \\ & truck overtaken and left behind on the left side of the bus, while a green motorcycle zoomed past on \\ & the opposite lane. \\
    \midrule
    \multirow{3}{*}{\textbf{Query}} & Objects: red sandwich, yellow carrot, brown hot dog, green cake \\
    & Relation: yellow carrot right of green cake \\
    & Sentence: \\
    \bottomrule
    \end{tabular}
    }
    \caption{Complete instruction and example demonstration used to generate the GPT-synthetic Dataset.}
    \vspace*{-0.08in}
    \label{tab:prompt-demo}
\end{table*}

\section{Implementation Details}\label{Append:implement-detail}

To help reproduce our results, we include a  comprehensive report of hyperparameters and the model architectures used in this work in Table~\ref{tab:supp-hyperparameters}.

The \texttt{Roberta-base} encoder \cite{liu2019roberta} in layout predictor consists of 12 layers and 12 heads, with a hidden dimension size of 768. We fine-tune the model on GPT-synthetic dataset with relative position objective and MS-COCO dataset with absolute position objective for 100 epochs. The training batch size is 64, and the learning rate of encoder starts at 1e-6 and decays to 1e-8. The learning rate of the GMM output layer starts at 4e-5 and decays to 1e-8. We use the pre-trained \texttt{ViT-B/32}~\cite{radford-clip-2021} to calculate CLIP similarities.
For the diffusion model, we adopt the pre-trained \texttt{stable-diffusion-v1-4}~\cite{rombach2022high} and stick with the default parameters. 
When optimizing the combination weights of cross-attention, the initial value is set to \e{1/N}, where \e{N} is the number of objects. The weight is projected to \e{[-1, 2]} after each gradient descent step to avoid extreme values.

\section{Details of GPT-synthetic Dataset and Dataset Statistics}\label{Append:datasets}

In this section, we detail the process of creating the GPT-synthetic dataset and report the statistics of each dataset.

The GPT-synthetic dataset contains the 80 object categories in MS-COCO \cite{lin2014microsoft}, and each description contains 2-5 objects and 1-4 relations.
To create a text description with \e{N} objects and \e{M} relations, \e{N} objects are first sampled without replacement from the same MS-COCO super-category (\emph{e.g.,} \e{N} objects from furniture), so that they are more likely to appear together in the same scene in real world. A color attribute is randomly assigned to each object with probability 0.5, and the assigned color is randomly sampled from a pre-defined list of colors.
Among the \e{N} objects, \e{M} pairs are then sampled without replacement and randomly assigned a spatial relation from ``left of,'' ``right of,'' ``above,'' and ``below.'' We consider these four relations because they can be easily and reliably measured by comparing the center position, and we additionally check the relations to ensure no contradiction exists (\emph{e.g.,} A is above B, B is above C, and C is above A). With specified objects and relations, GPT3~\cite{gpt3} is prompted to generate a sentence that mentions all objects and relations, given 5 demonstration examples. We specifically instruct GPT3 to generate diverse sentences. Table~\ref{tab:prompt-demo} shows the complete instruction, a sample demonstration, and a query that is used to generate a sentence. In practice, we manually write 20 demonstrations and randomly sample 5 for each generation.

Table~\ref{tab:append-data-stats} shows the number of text descriptions for the GPT-synthetic dataset.
For the other two datasets, VSR contains 500 descriptions, and each description involves two objects and one spatial relation. MS-COCO contains 500 descriptions, including 200 descriptions with two objects, 150 descriptions with three objects, and 150 descriptions with four objects. There are no explicit spatial relations in MS-COCO descriptions. In general, GPT-synthetic contains the most complex text descriptions in terms of the number of objects and spatial relations.

\section{CLIP Similarity}\label{Append:CLIP}

We additionally use CLIP similarity~\cite{radford-clip-2021} to measure how well the generated image aligns with the text description.
Specifically, we consider CLIP similarity at two different granularities. \textbf{Global CLIP score} calculates the CLIP similarity between the whole text description and the whole image. On the other hand, \textbf{local CLIP score} calculates the similarity between the local object description and its corresponding bounding box in the image, if the object is detected. We use the local description defined in Sec.~\ref{sec:method-overview}, \emph{e.g.,} ``A photo of a black mailbox.''

The performance is presented in Table~\ref{tab:CLIP_similarity}. We observe that different methods have very close global and local CLIP scores, which may be ascribed to the limited capability of vision-and-language models when text descriptions consist of multiple objects [59]. As such, the CLIP similarity is not sufficient to indicate which method performs better in our experiments, and we leverage the subjective evaluation in Sec. \ref{exp} for a more informative analysis.

\section{Layout Predictor Analyses}\label{Append:layout-effect}
In this section, we analyze how the layout predictor affects our method from the following two aspects.

\vspace*{0.05in}
\noindent
\textbf{Layout predictor helps synthesize correct spatial relations} \quad
We first demonstrate the position of each object predicted by our layout predictor and further show how the predicted position helps diffusion models generate objects with correct spatial relations. 
Fig.~\ref{fig:layout-predictor-help} illustrates two examples. Given the text description, the first column demonstrates images synthesized directly by vanilla stable diffusion model, where some objects are mislocated (\emph{e.g.,} the spoon and bowl) and missed (\emph{e.g.,} sandwich).
The second column shows the predicted position of each object by our layout predictor, where it correctly locates objects according to the specified spatial relations (\emph{e.g.,} the apple is placed beneath the sandwich).
Finally, the last column shows images can be generated following the predicted layout, thus locating objects at correct position.

\begin{figure}
\centering
\includegraphics[width=0.4\textwidth]{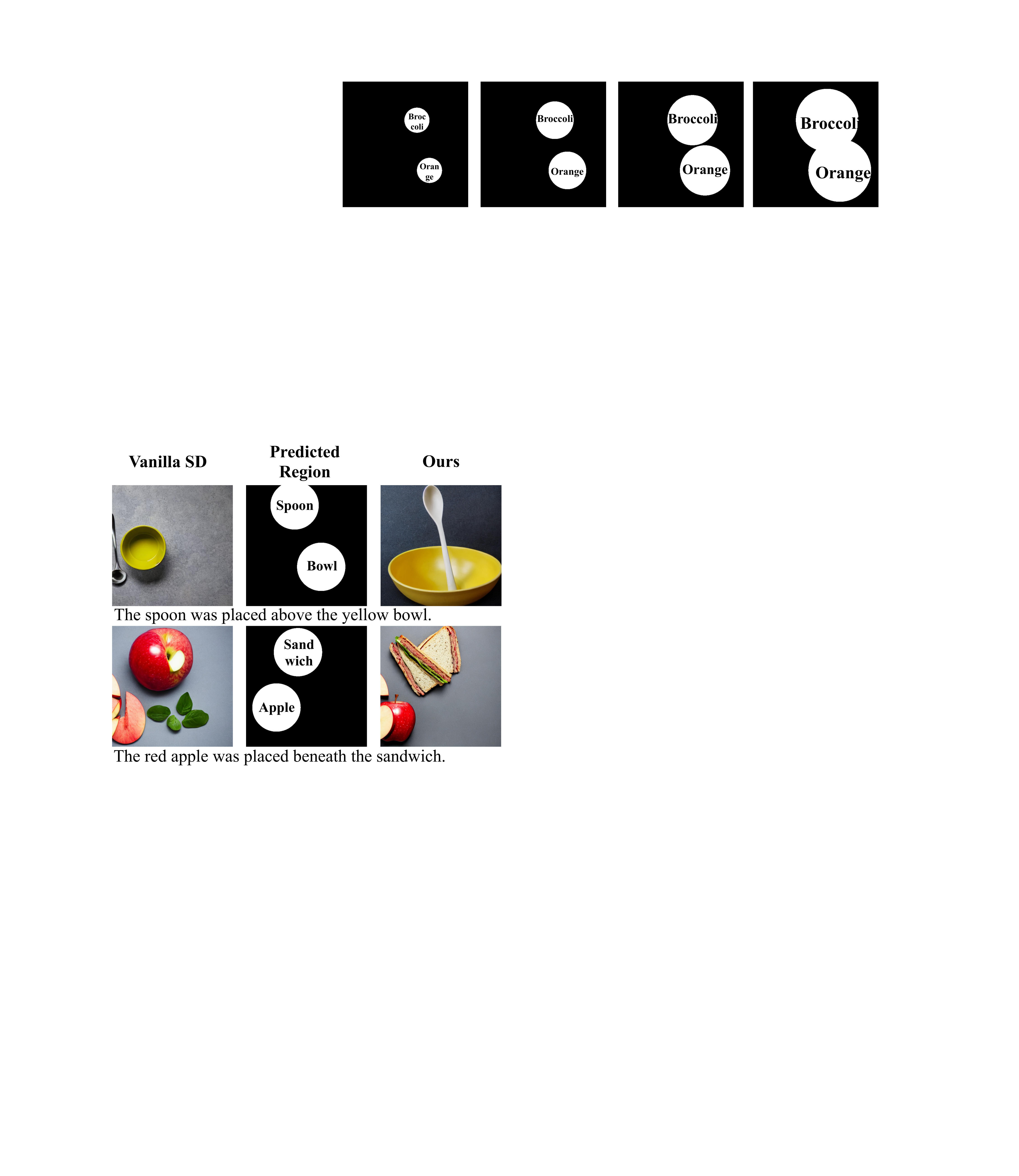}
\caption{Layout predictor helps generate objects at correct locations. First column: Images generated by vanilla stable diffusion model. Second column: Pixel region generated by our layout predictor. Third column: Images generated by out method.}
\vspace*{-0.1in}
\label{fig:layout-predictor-help}
\end{figure}

\begin{figure}
\centering
\includegraphics[width=0.45\textwidth]{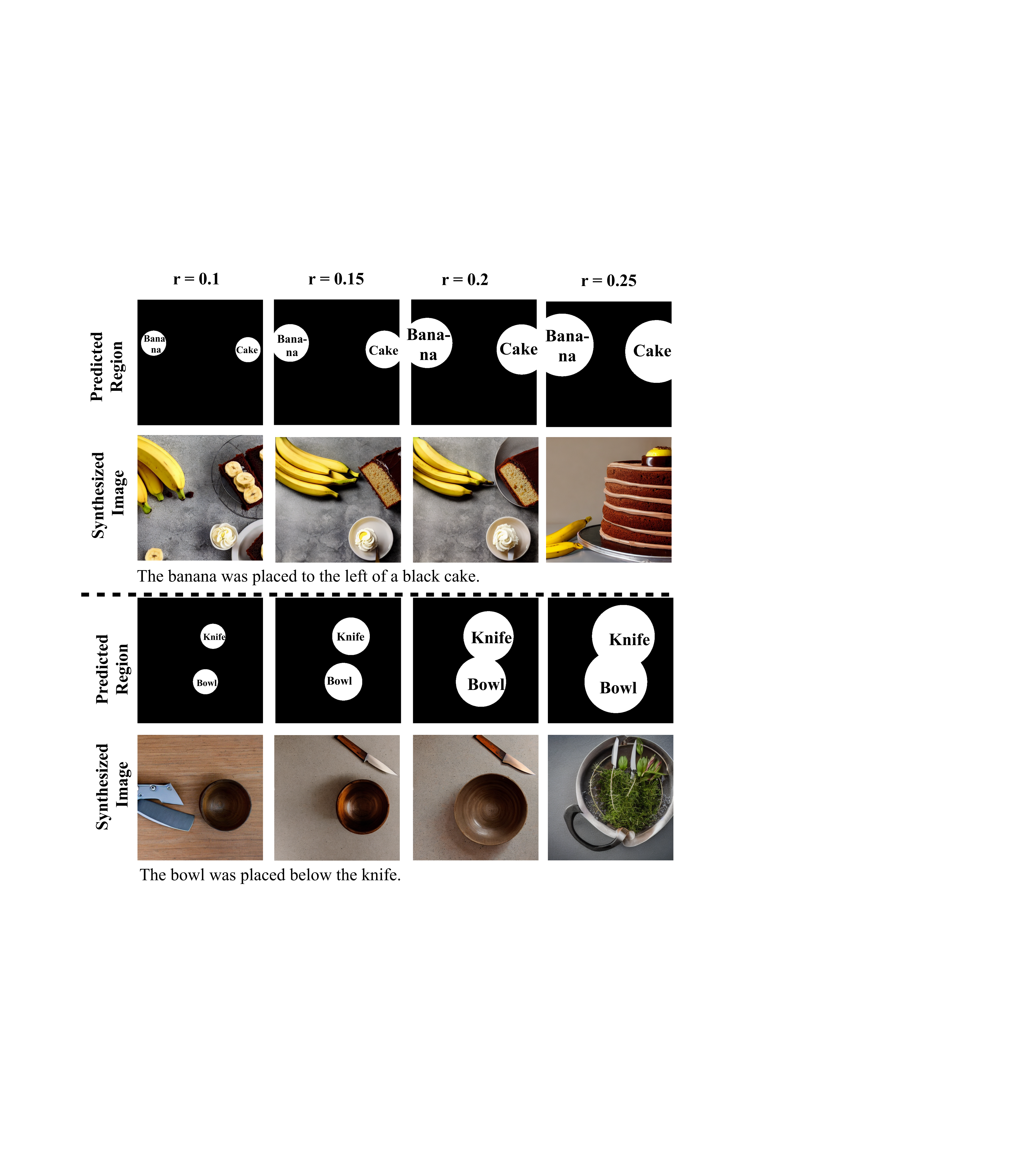}
\caption{Example images generated by our method with different radii. For each panel, the top row is the predicted regions with different radii, and the bottom row is the generated images.}
\vspace*{-0.05in}
\label{fig:radius-variance}
\end{figure}

\vspace*{0.05in}
\noindent
\textbf{Layout predictor does not restrict the object size} \quad
Once the position of each object is predicted, we define the pixel region of each object as a circle centered at the predicted coordinate with a radius \e{r}.
We demonstrate two examples for images generated with different radii \e{r} in Fig.~\ref{fig:radius-variance}.
Generally, the object size increases as the radius increases.
However, the circular region does not strictly bound the object, and objects can go beyond the region (\emph{e.g.,} the cake in the first row is larger than predicted region).
Moreover, the generation results are not highly sensitive to the choice of \e{r}, as shown by the similar outputs of \e{r=0.15} and \e{r=0.2}. We thus fix \e{r=0.2} in our experiments.

\section{Ablation Study for Layout Predictor }\label{Append:ablation-layout}

\noindent
\textbf{Using ground truth location in place of layout predictor} \quad
In Sec. \ref{sec:main-results}, we report the results based on our layout predictor. We now investigate the performance when ground truth location is used in place of the layout predictor. Specifically, we use the center of the ground truth bounding box as the object position, and we use the same radius \e{r=0.2} to construct the pixel region. We evaluate the performance on VSR dataset since it contains the ground truth bounding box information. As can be observed in Table~\ref{tab:layout-ablation}, providing ground truth location boosts the performance, especially for spatial relation precision, since the pixel region is guaranteed to preserve the correct spatial relations.
We also evaluate the performance of \textsc{Paint-with-Words} baseline when the ground truth position is given. It achieves 64.7\% object recall and 58.3\% SPRel precision, which is worse than the counterpart of our method.

\vspace*{0.05in}
\noindent
\textbf{Training layout predictor with only absolute or relative position objective} \quad
Our layout predictor is jointly trained with both absolution and relative position objectives (Sec.~\ref{sec:layout-predictor}). 
We now explore our method's performance when the predictor is trained with only one of the objectives. The results are shown in Table~\ref{tab:layout-ablation}. We observe that both settings lead to performance drop, indicating that both objectives are critical for an effective layout predictor. Moreover, removing relative position objective leads to significant performance degradation on SPRel precision, which demonstrates the importance of the objective.

\begin{table}
\centering
\resizebox{0.9\linewidth}{!}{%
\begin{tabular}{lcc}
\toprule
  & Object Recall  & SPRel Precision \\

\midrule
Ours & 65.1\% & 75.0\%\\
\midrule
Ground truth position & 66.3\% & 78.1\%\\
No absolute position obj &62.7\% & 68.2\%  \\
No relative position obj & 63.5\% & 58.6\%\\
Soft pixel region & 64.3\% & 69.8\%\\
\bottomrule
\end{tabular}
}
\caption{Ablation study for layout predictor on VSR dataset.}
\label{tab:layout-ablation}
\vspace*{-0.08in}
\end{table}

\vspace*{0.05in}
\noindent
\textbf{Hard versus Soft Threshold on Pixel Region} \quad
We use a hard threshold to get the pixel region in Sec.~\ref{sec:spatial-temporal}. Here we explore a different strategy that produces a soft pixel region for an object.
Specifically, we expand the pixel region of an object to the whole image but assign a smaller weight for pixels that are further away from the object. Formally, the output of the attention layer at time \e{t} becomes
\begin{equation}
\vspace*{-0.05in}
\small
\begin{aligned}
    \bm O(t) &= \sum_{i=1}^N \lambda_{it} \bm G_i \odot \textrm{Attention}(\bm Q, \bm K_{\bm L_i}, \bm V_{\bm L_i}) \\
    &+ \Big(1 - \sum_{i=1}^N \lambda_{it} \bm G_i\Big) \odot \textrm{Attention}(\bm Q, \bm K_{\bm D}, \bm V_{\bm D}),
\end{aligned}
\label{eq:optimizer}
\end{equation}
where \e{\lambda_{it}, \bm Q, \bm K_{\bm L_i}, \bm V_{\bm L_i}, \bm K_{\bm D}, \bm V_{\bm D}} are defined in Sec.~\ref{sec:spatial-temporal}, and \e{\bm G_i} is a soft pixel region matrix for object \e{O_i}, with \e{\bm G_i(x, y) = g\left((x, y); \bm C_i, \sigma^2 \bm I\right) / g\left(\bm C_i; \bm C_i, \sigma^2 \bm I\right)}, where \e{g\left((x, y); \bm C_i, \sigma^2 \bm I\right)} is the probability density of a 2D Gaussian distribution with mean \e{\bm C_i} and covariance matrix \e{\sigma^2 \bm I} at point \e{(x, y)}, \e{\bm C_i} is the center coordinate of the object, and \e{\sigma} is a hyperparameter.
Intuitively, the combination weight of an object decreases as the pixel moves away from the object center, and the weights are normalized so that the object center has combination weight \e{\lambda_{it}}. The performance of this strategy is shown in Table~\ref{tab:layout-ablation}, where it achieves a slightly worse performance compared to the hard threshold region.

Finally, we show examples in Fig.~\ref{fig:user-provide} where user provided region (possibly irregular) is given to our method. The generated images largely follow the provided layout, which demonstrates that our method can be adapted for image generation with better user interaction.

\begin{figure}
\centering
\includegraphics[width=0.45\textwidth]{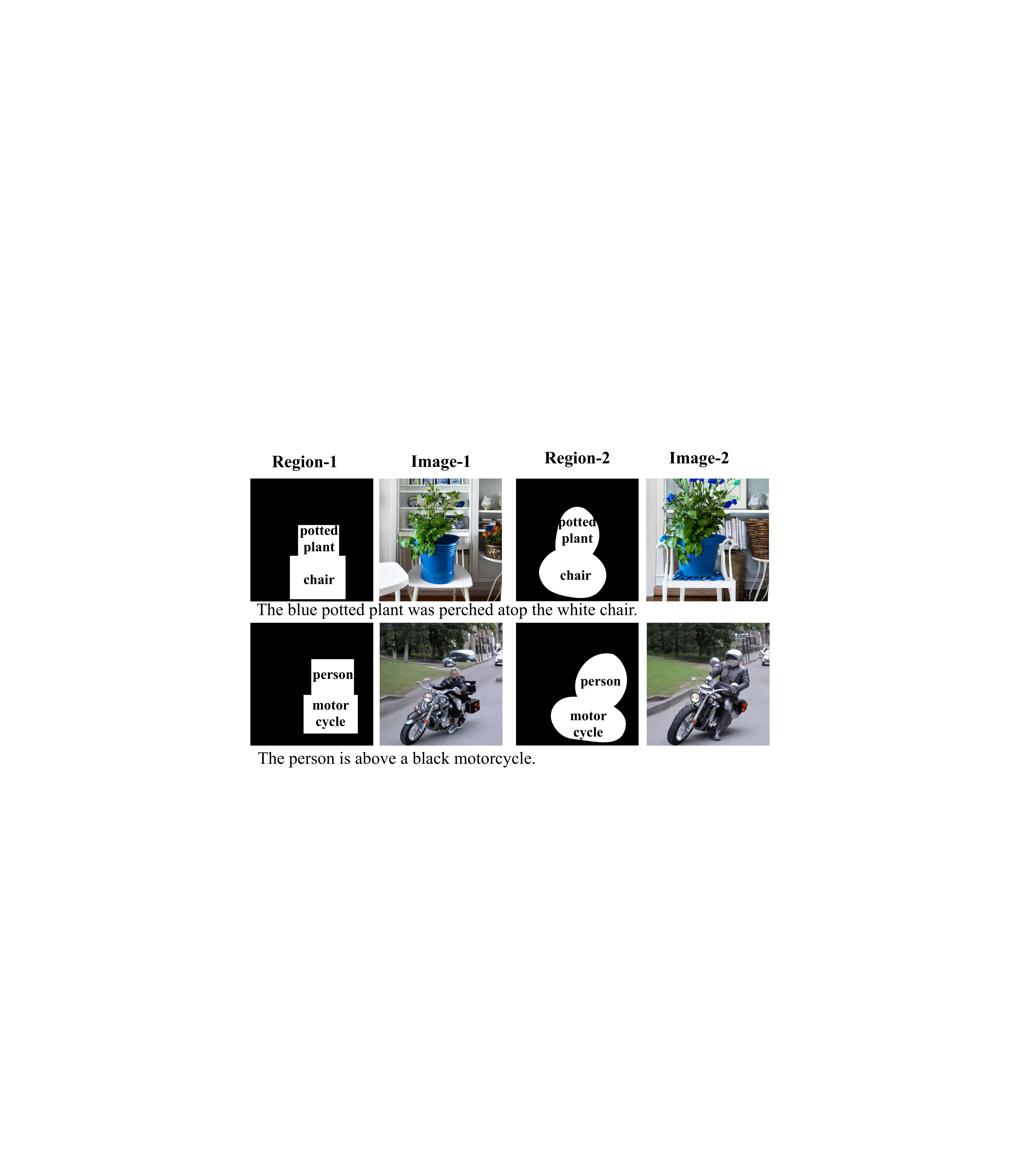}
\caption{Example images generated from user provided region. The regions are shown in the first and third columns, and corresponding images are shown in the second and fourth columns.}
\vspace*{-0.05in}
\label{fig:user-provide}
\end{figure}

\section{Performance on Uncommon Combinations}\label{Append:rare-objects}
To further test if our method can generate high-fidelity images for novel text descriptions, we demonstrate the performance of our method and baselines on a dataset that contains uncommon scenes. This uncommon synthetic dataset consists of 100 text descriptions, and it differs from the GPT-synthetic dataset in Sec.~\ref{sec:layout-predictor} from two aspects.
(1) When sampling objects for a description, we remove the constraint that objects need to belong to the same super category. Sampling without this constraint can thus produce rare object pairs (\emph{e.g.,} objects from food and vehicle can occur in the same description). (2) We manually check generated samples and only keep the ones that are unlikely to appear in real life.
The result is demonstrated in Table~\ref{tab:gpt-rare-res}. We observe that our method achieves the best result in terms of object recall and spatial relation precision, indicating that our method can better generalize to novel text descriptions. Some visual examples can be found in Fig.~\ref{fig:rare-combination} and Fig.~\ref{fig:uncommon-suppl}.

\begin{figure}[h]
\centering
\includegraphics[width=0.45\textwidth]{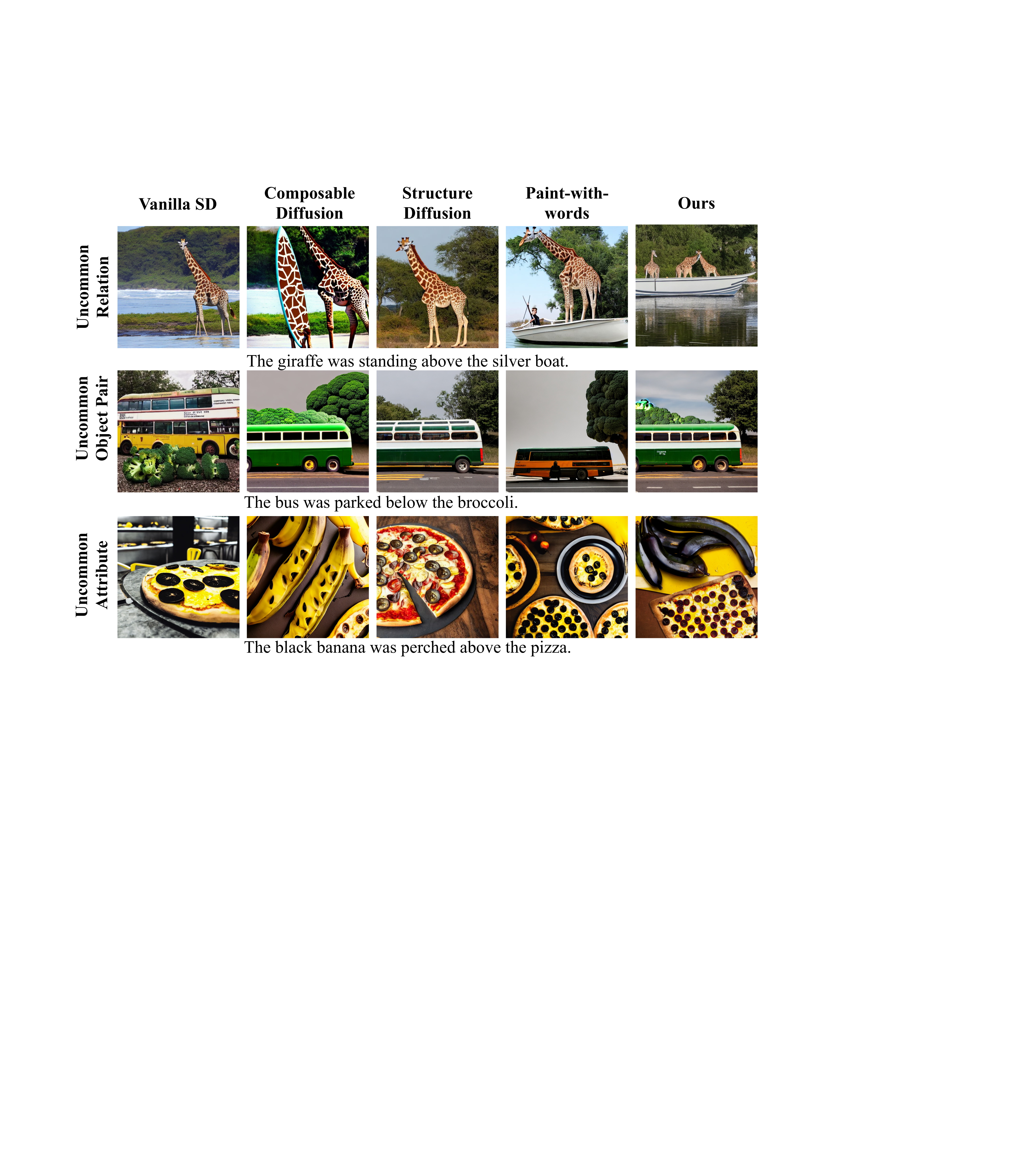}
\caption{Example images generated by our method and baselines on uncommon relations, object pairs, and attributes.}
\label{fig:uncommon-suppl}
\vspace*{-0.05in}
\end{figure}

\begin{table}
\centering
\resizebox{0.9\linewidth}{!}{%
\begin{tabular}{lcc}
\toprule
  & Object Recall  & SPRel Precision \\

\midrule
\textsc{Vanilla-SD} & 39.8\% & 52.6\%  \\
\textsc{Composable-Diffusion} & 30.1\%&  50.8\%\\
\textsc{Structure-Diffusion} & 40.1\% & 51.6\% \\
\textsc{Paint-with-Words} & 41.2\% & 54.7\% \\
Ours & \textbf{42.4\%} & \textbf{59.6\%}\\

\bottomrule
\end{tabular}
}
\caption{Performance on text descriptions that contain uncommon object pairs, object-attribute pairs, and spatial relations.}
\label{tab:gpt-rare-res}
\vspace*{-0.05in}
\end{table}

\section{More Examples and Failure Cases}\label{Append:qualitative-examples}
In this section, we provide more example images from our method and baselines. Then, we analyze the potential failure cases of our method.

\vspace*{0.05in}
\noindent
\textbf{More Examples} \quad
We provide more example images from our method and baselines in Fig. \ref{fig:qualitative-analysis-more}. The results are consistent with Fig.~\ref{fig:qualitative-analysis}, where our method generates images with high object, attribute, and spatial fidelities.

\vspace*{0.05in}
\noindent
\textbf{Failure Cases}\label{Sec:Failure}
\quad
We present two failure cases in Fig.~\ref{fig:failure-case}. For each example, we first show the predicted pixel region by our layout predictor and the corresponding generated image (left two columns).
We observe that these predicted positions tend to be at the edge of the image, which reduces the region of the object. We hypothesize that this will lead to insufficient attention to the corresponding object, and thus the object cannot be successfully synthesized (\emph{e.g.,} the cell phone in the first row).
We further demonstrate in the right two columns that moving pixel regions inside the image can resolve these failure cases, where the missing objects can be synthesized.
Future work may consider adding the constraint that the predicted object center cannot locate at the edge of the image.

\begin{figure}[h]
\centering
\includegraphics[width=0.45\textwidth]{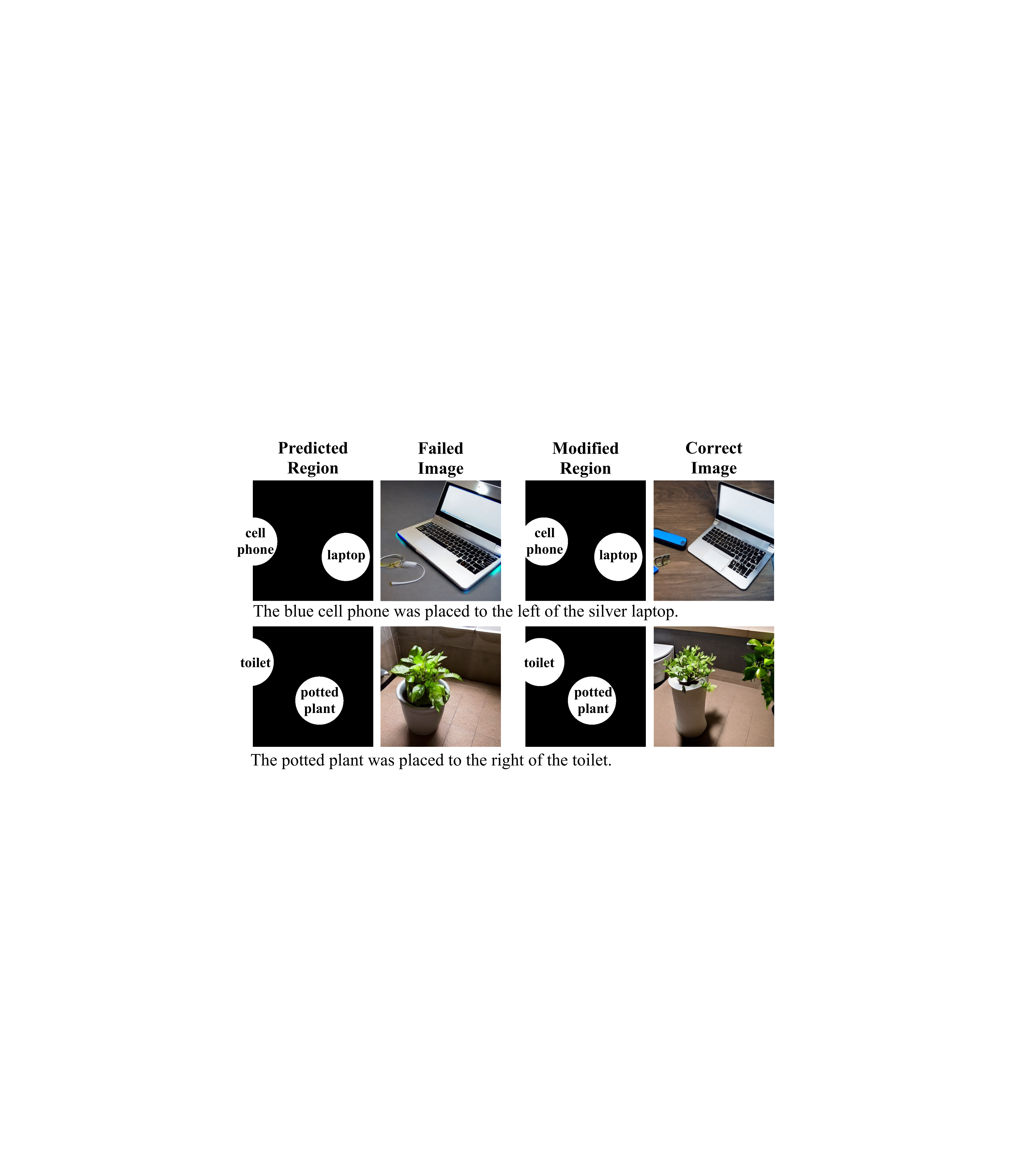}
\caption{Failure Cases. The first two columns show the predicted region and the synthesized image, where some objects are missing. The last two columns demonstrate that modifying the pixel region can resolve the problem.}
\label{fig:failure-case}
\vspace*{-0.05in}
\end{figure}

\begin{figure*}
\centering
\vspace{-0.04in}
\includegraphics[width=0.93\textwidth]{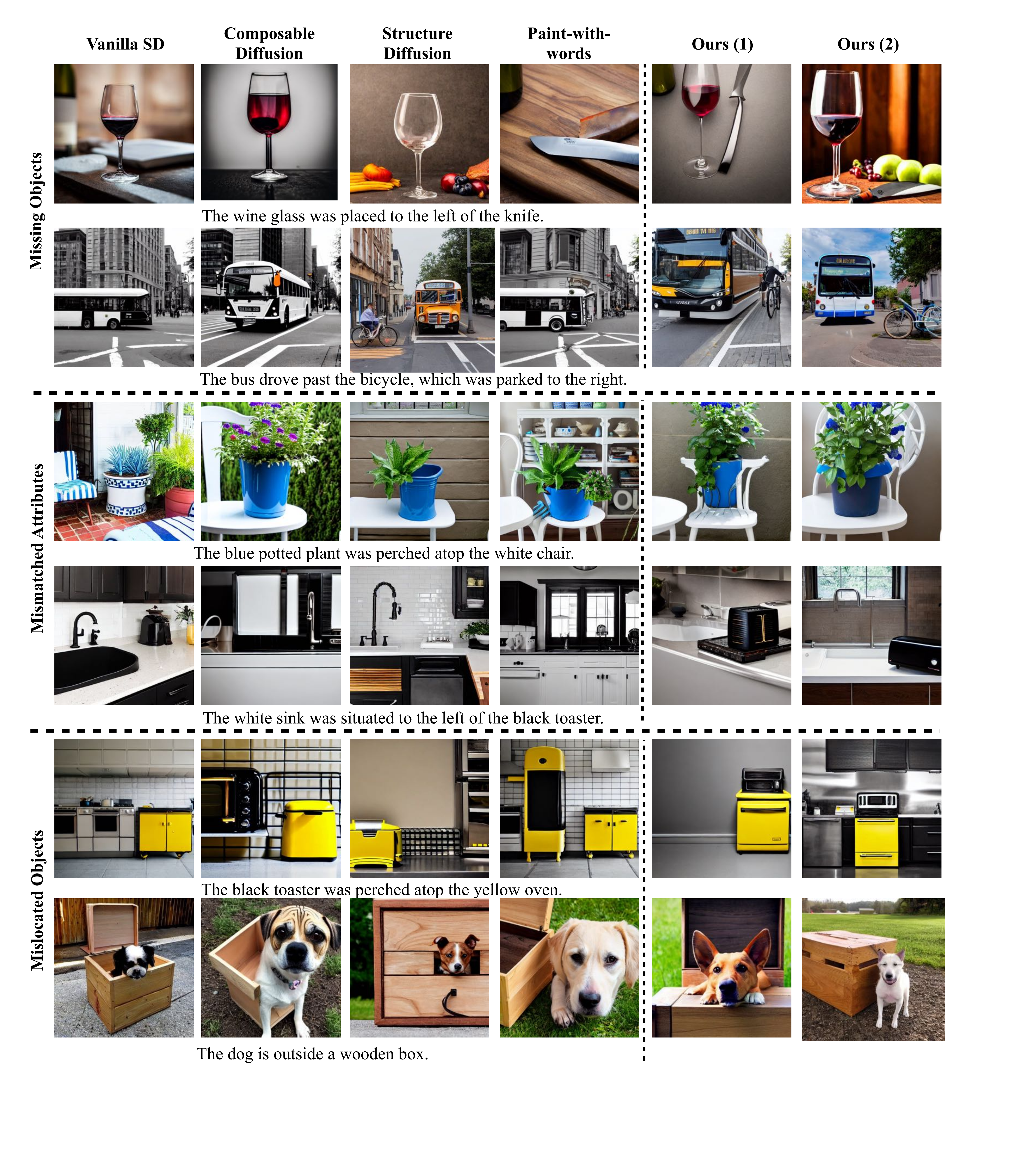}
\vspace{-0.04in}
\caption{\textbf{Example images generated by our method and baselines.} Typical errors of baselines include missing objects, mismatched attributes, and mislocated objects. Ours (1)/(2) show the results with two different random seeds.}
\vspace*{-0.05in}
\label{fig:qualitative-analysis-more}
\end{figure*}

\section{Details of Subjective Evaluation}\label{Append:subjective-eval}
In this section, we provide further details of our subjective evaluation.
We compare with all baselines on MS-COCO, VSR, and GPT-synthetic datasets.
For each dataset, we randomly sample 25 text descriptions for evaluation. We evaluate on Amazon Mechanical Turk, and 85 workers participate in our study. During the subjective evaluation, the workers are asked four questions: (1) (\textit{Object Fidelity}) Does the image contain all objects mentioned in the text? (2) (\textit{Attribute Fidelity}) Are all synthesized objects consistent with their characteristics described in the text (\emph{e.g.,} color and material)? (3) (\textit{Spatial Fidelity}) Does the image locate all objects at the correct position such that the spatial relations in the text are satisfied (if an object in the relationship is missing, it is considered as an incorrect generation)? and (4) (\textit{Overall}) Which image in the pair has higher fidelity with the text and has better quality?
For the first three questions, we present the participant with a single image generated by one method, and ask the participant to rate the image using a score of 0, 1, or 2, where 2 denotes all objects/attributes/relations are correct and 0 denotes none of them is correct. For the last question, participant will see a pair of images, where one of them is generated by our method and the other one is generated by one baseline. The participant is then asked to select the better image in terms of overall fidelity and quality. The subjective evaluation interface is shown in Fig.~\ref{fig:amt_instructions_1} and Fig.~\ref{fig:amt_instructions_2}.
The subjective evaluation results are shown in Table~\ref{tab:subjective_eval}. We also provide all generated images by our method and baselines in Figures \ref{fig:supp-subjective-vsr}, \ref{fig:supp-subjective-gpt}, and \ref{fig:supp-subjective-mscoco}.

\paragraph{References} \text{ } 
\\
\text{[59]} Tristan Thrush, Ryan Jiang, Max Bartolo, AmanpreetSingh, Adina Williams, Douwe Kiela, and Candace Ross. Winoground: Probing vision and language models for visio-linguistic compositionality. In \textit{CVPR}, 2022.

\begin{table*}[t]
    \small
	\begin{tabular}{p{1\linewidth}}
	\hline
        \textbf{Instructions:}\\
		\arrayrulecolor{black}  %
		\midrule
		Please read the instructions carefully. Failure to follow the instructions may lead to rejection of your results. Your task will involve evaluating whether target objects have been successfully synthesized using AI models. First, you will see a text description that outlines the objects to be generated (e.g., “The bed is below the black cat.”). Then you will see an image, which is generated based on the provided text by an AI algorithm. You will then be asked to evaluate if the generated image contains all the objects mentioned in text. You will use a scoring system ranging from 0 to 2, where 0 indicates all objects are incorrect or missing, 1 means some objects are incorrect or missing, and 2 means all objects are successfully generated. Notice that you should \textbf{only} rate if the objects are synthesized or not; you should disregard their inconsistencies with text description such as colors or relative positions (e.g., left/right).\\
    \vspace*{-0.05 in}\textbf{Example:} We provide an example to help you understand how to evaluate the generated results. The text description is ``\textbf{The bed is below the black cat.}'' \\

    \begin{center}
    \includegraphics[width=0.70\textwidth]{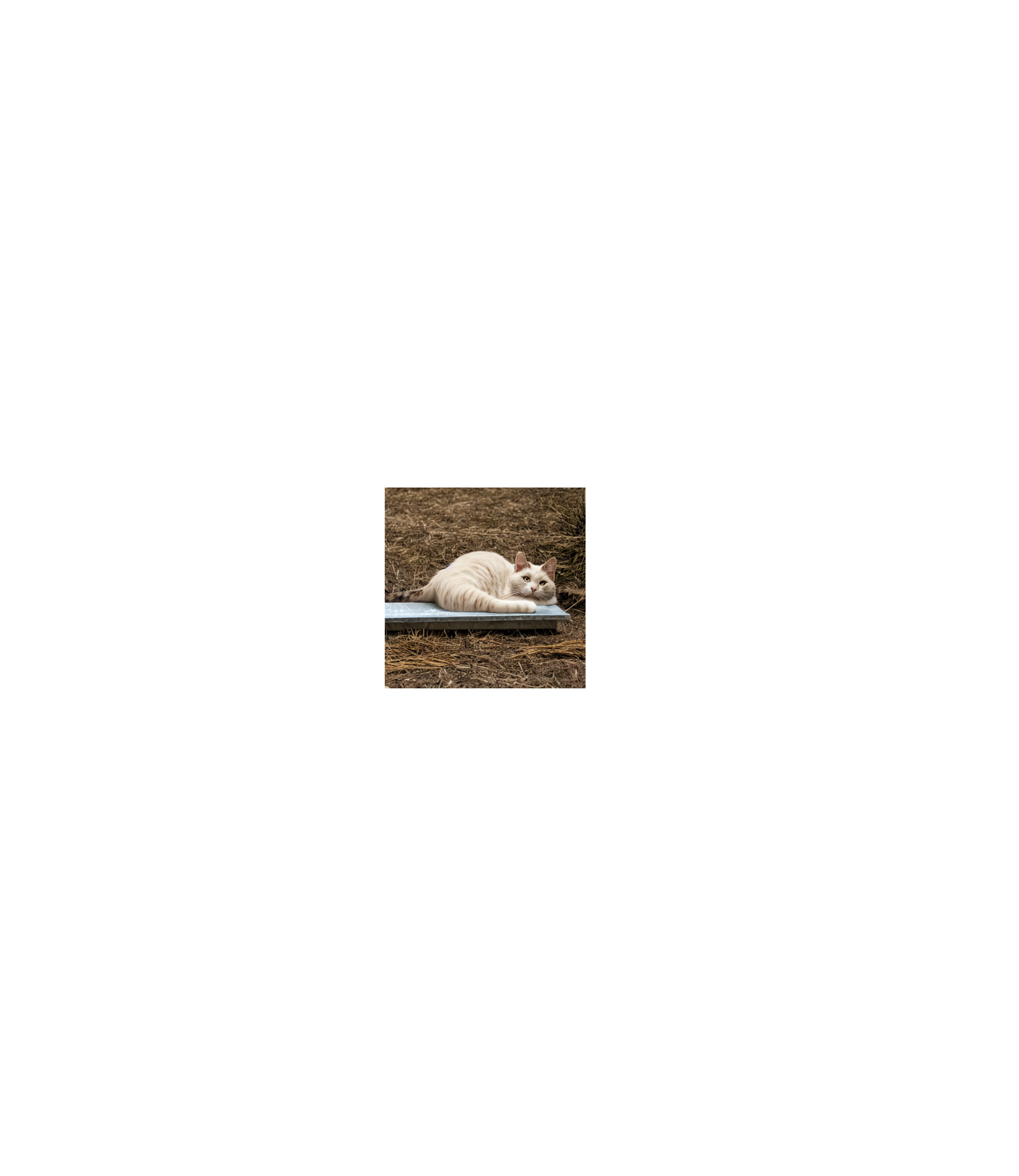}
    \end{center}
    
    \vspace*{-0.1in} We can observe that cat is successfully generated, but the bed is not. Therefore, this example is partially correct, and you should rate score 1. Again, notice that it synthesizes a white cat while the text says “black cat”, but you should ignore the inconsistencies of color and position. \\
    
    \arrayrulecolor{black}  %
	\midrule
    \vspace*{-0.14 in}\textbf{Question:} \\
    \midrule
    \vspace*{-0.13 in} The text description is ``\textbf{The motorcycle is parking to the right of a bus.}'' Does this image contain the objects mentioned in the text description? Rate the generation results from 0 (all objects missed) to 2 (all objects are generated).\\
    \begin{center}
    \includegraphics[width=0.70\textwidth]{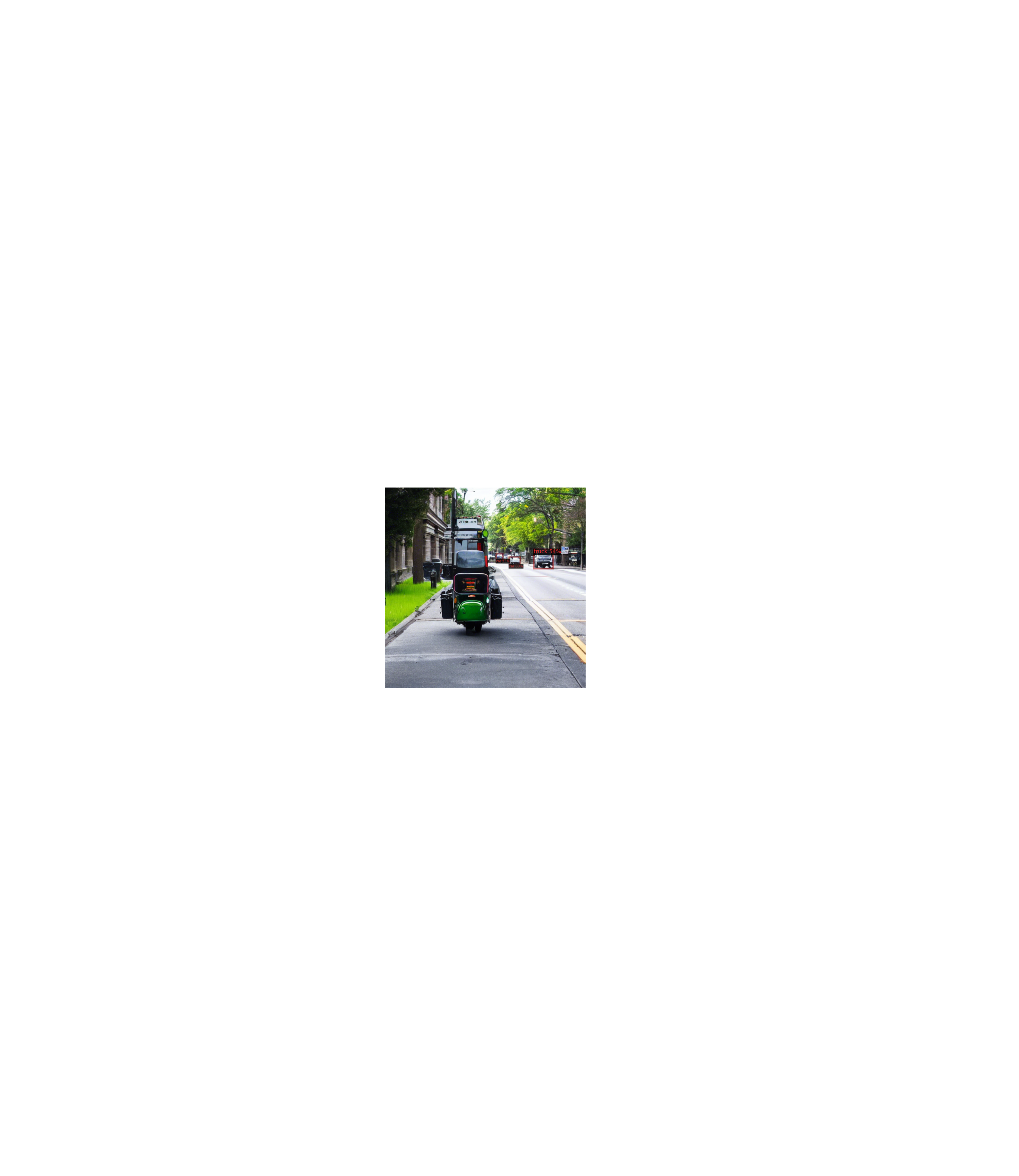}
    \end{center}
 
    \vspace*{-0.1 in}
     $\square$ 0 \\
     $\square$ 1 \\
     $\square$ 2 \\
\hline
	\end{tabular}
    \captionof{figure}{\small{Instructions and an example question of the subjective evaluation on Amazon Mechanical Turk. The goal is to evaluate whether the generated images contain all specified objects (object fidelity). The interfaces for attribute fidelity and spatial fidelity are similar.}}
    \label{fig:amt_instructions_1}
\end{table*}

\begin{table*}[t]
    \small
	\begin{tabular}{p{1\linewidth}}
	\hline
        \textbf{Instructions:}\\
		\arrayrulecolor{black}  %
		\midrule
		Please read the instructions carefully. Failure to follow the instructions will lead to the rejection of your results. In this task, you will be asked to judge and compare the quality of two AI-generated images. Specifically, you will first see a text description, which describes the desired content we want to generate (e.g., “The bed is below the white cat.”). Then you will see two images, which are generated based on the provided text by different AI algorithms. You will then be asked to evaluate which image better follows the text description. When evaluating, you should consider the following aspects: (1) Does the synthesized image contain all objects mentioned in the text? (2) Does each object in the image follow the text description? (3) Does the image preserve the correct spatial relations mentioned in the text? (4) Does the image look real and natural? Then, you will choose the better image from the two candidate images.\\
    \vspace*{-0.05 in}\textbf{Example:} We provide an example to help you understand how to evaluate the generated results. The text description is ``\textbf{The bed is below the white cat.}'' \\

    \begin{center}
    \includegraphics[width=0.60\textwidth]{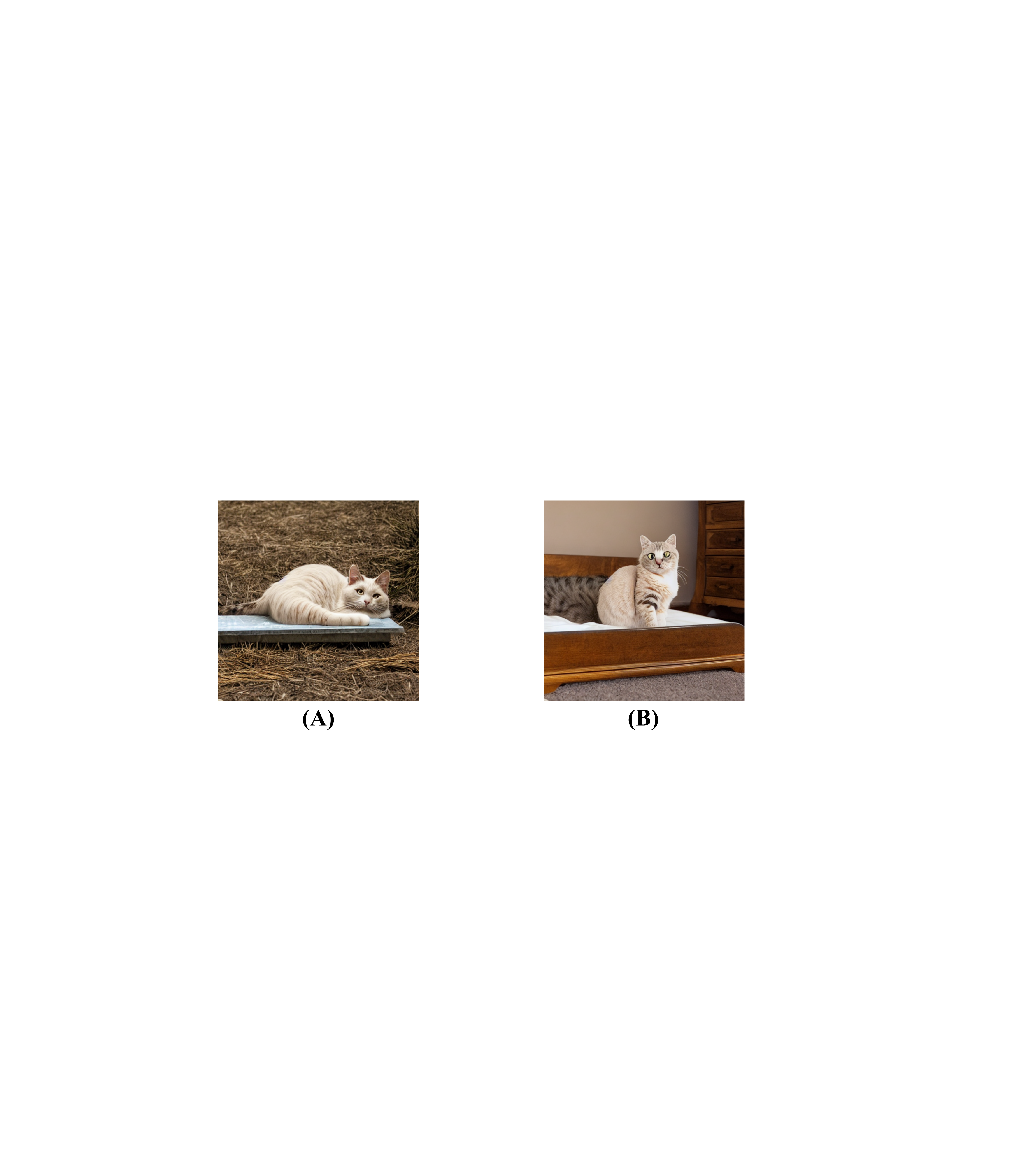}
    \end{center}
    
    \vspace*{-0.1in}You will evaluate based on the above criteria. First, it is important that the edited images faithfully synthesize the two objects “bed” and “cat” in the image.  All methods generate a cat in the image. However, method A fails to generate high quality “bed”, while method B generates a better bed. Second, both methods try to generate a white-colored cat. Third, both methods preserve the correct spatial relation that the cat is above a bed. Finally, the cat in method B looks more natural. Considering all the above analysis, method B is better. \\
    
    \arrayrulecolor{black}  %
	\midrule
    \vspace*{-0.14 in}\textbf{Question:} \\
    \midrule
    \vspace*{-0.13 in} The text description is ``\textbf{The motorcycle is parking to the right of a bus.}'' Which image in the pair has higher fidelity with the text and has better quality? Please give an overall evaluation based on the above criteria.\\
    \begin{center}
    \includegraphics[width=0.60\textwidth]{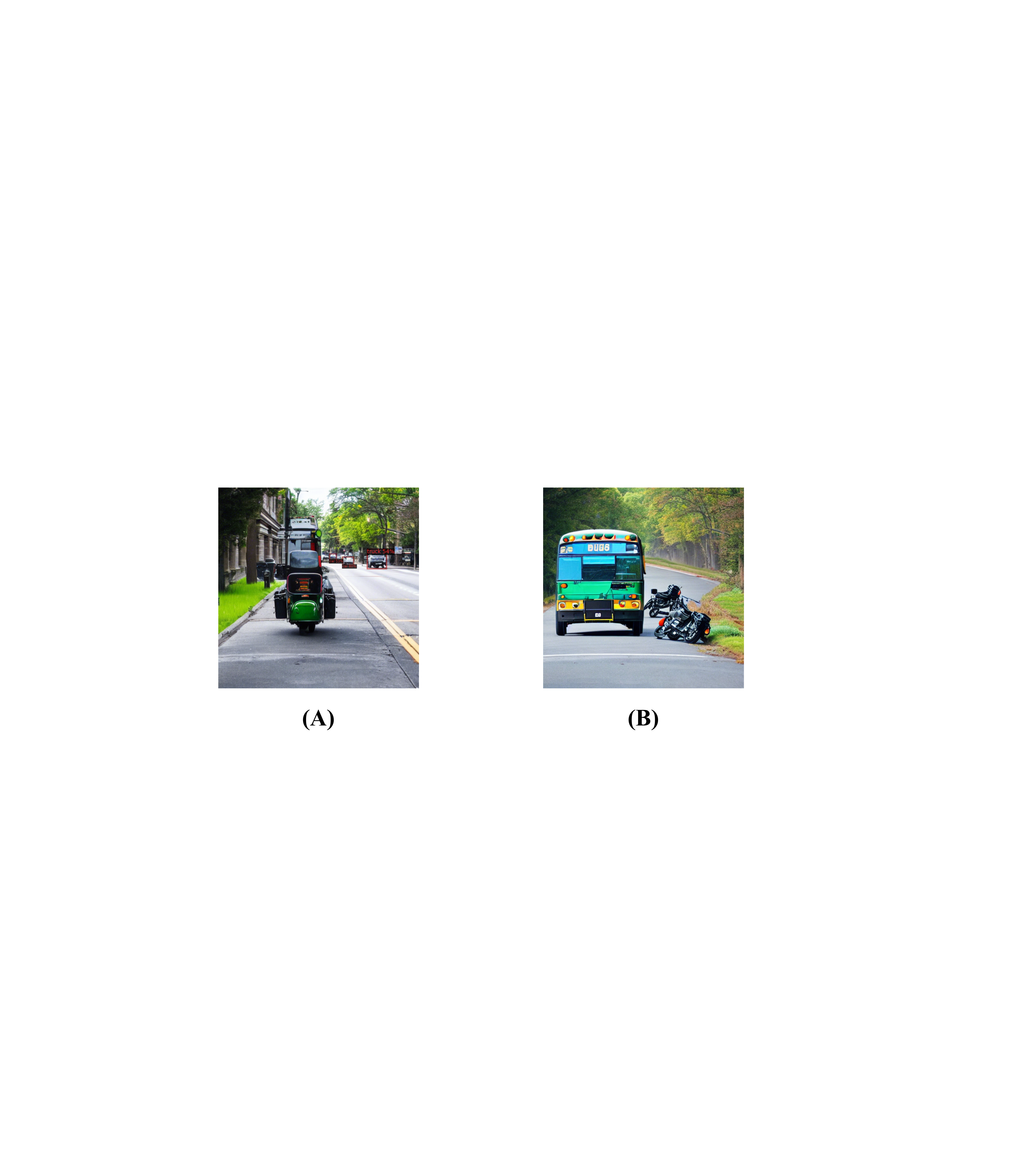}
    \end{center}
 
    \vspace*{-0.1 in}
     $\square$ (A) \\
     $\square$ (B) \\
\hline
	\end{tabular}
    \captionof{figure}{\small{Instructions and an example question of the subjective evaluation on Amazon Mechanical Turk. The goal is to compare two images generated by baselines and our method.}}
    \label{fig:amt_instructions_2}
\end{table*}

\begin{figure*}
\centering
\vspace{-2.0em}
\includegraphics[width=0.90\textwidth]{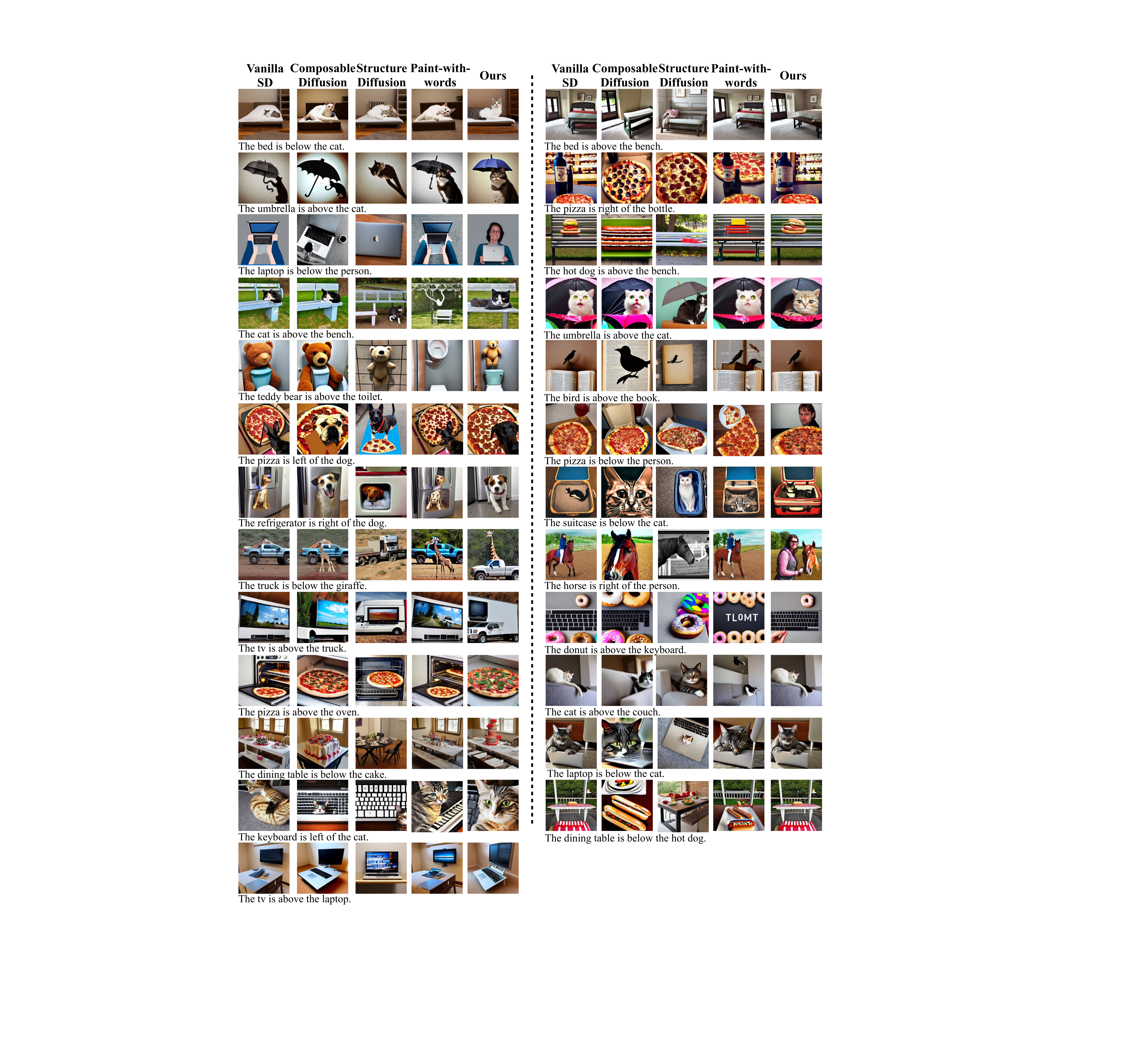}
\caption{Generated images for subjective evaluation on VSR dataset.}
\vspace*{-0.1in}
\label{fig:supp-subjective-vsr}
\end{figure*}

\begin{figure*}
\centering
\vspace{-2.0em}
\includegraphics[width=0.90\textwidth]{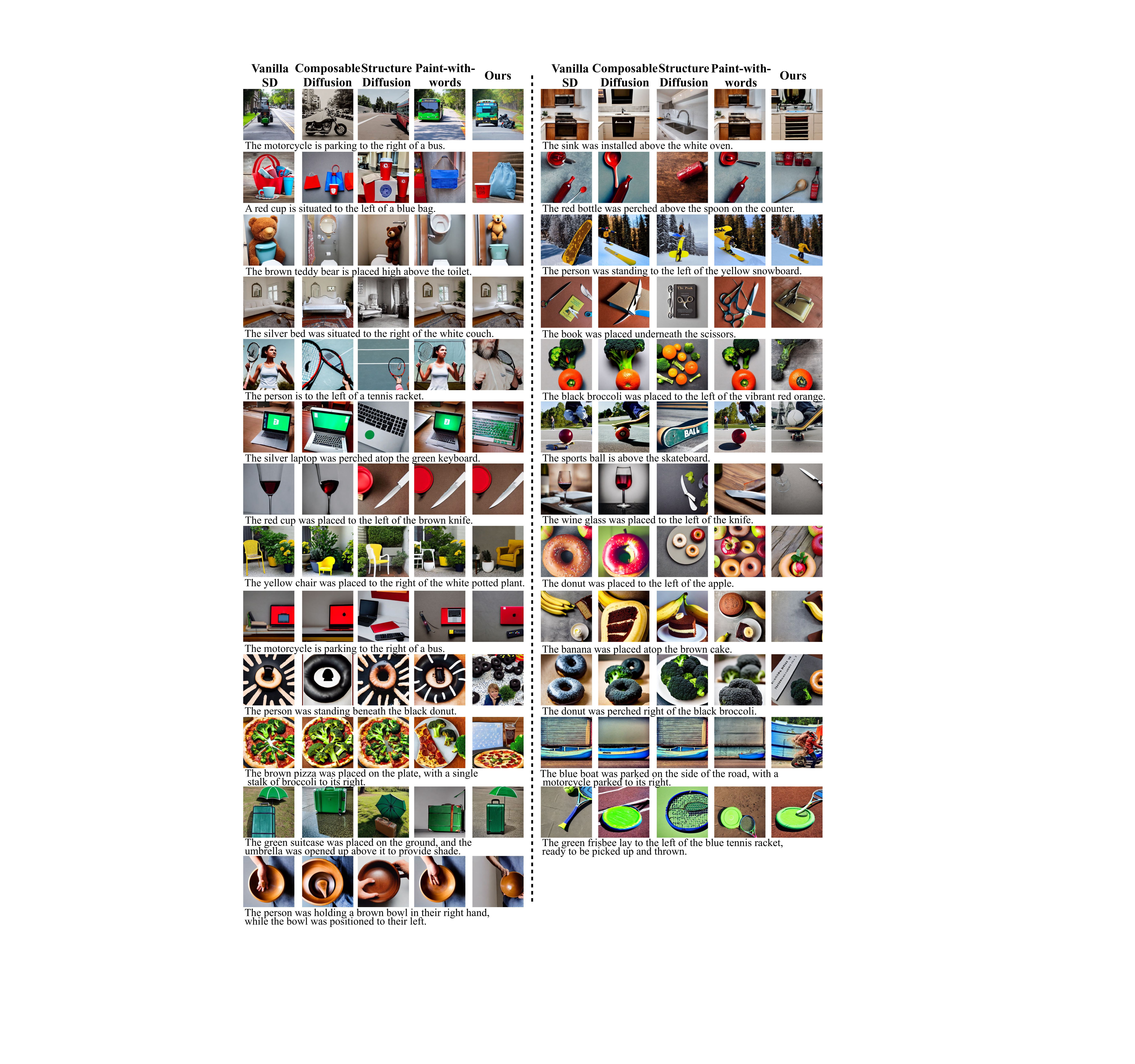}
\caption{Generated images for subjective evaluation on GPT-synthetic dataset.}
\vspace*{-0.1in}
\label{fig:supp-subjective-gpt}
\end{figure*}

\begin{figure*}
\centering
\vspace{-2.0em}
\includegraphics[width=0.90\textwidth]{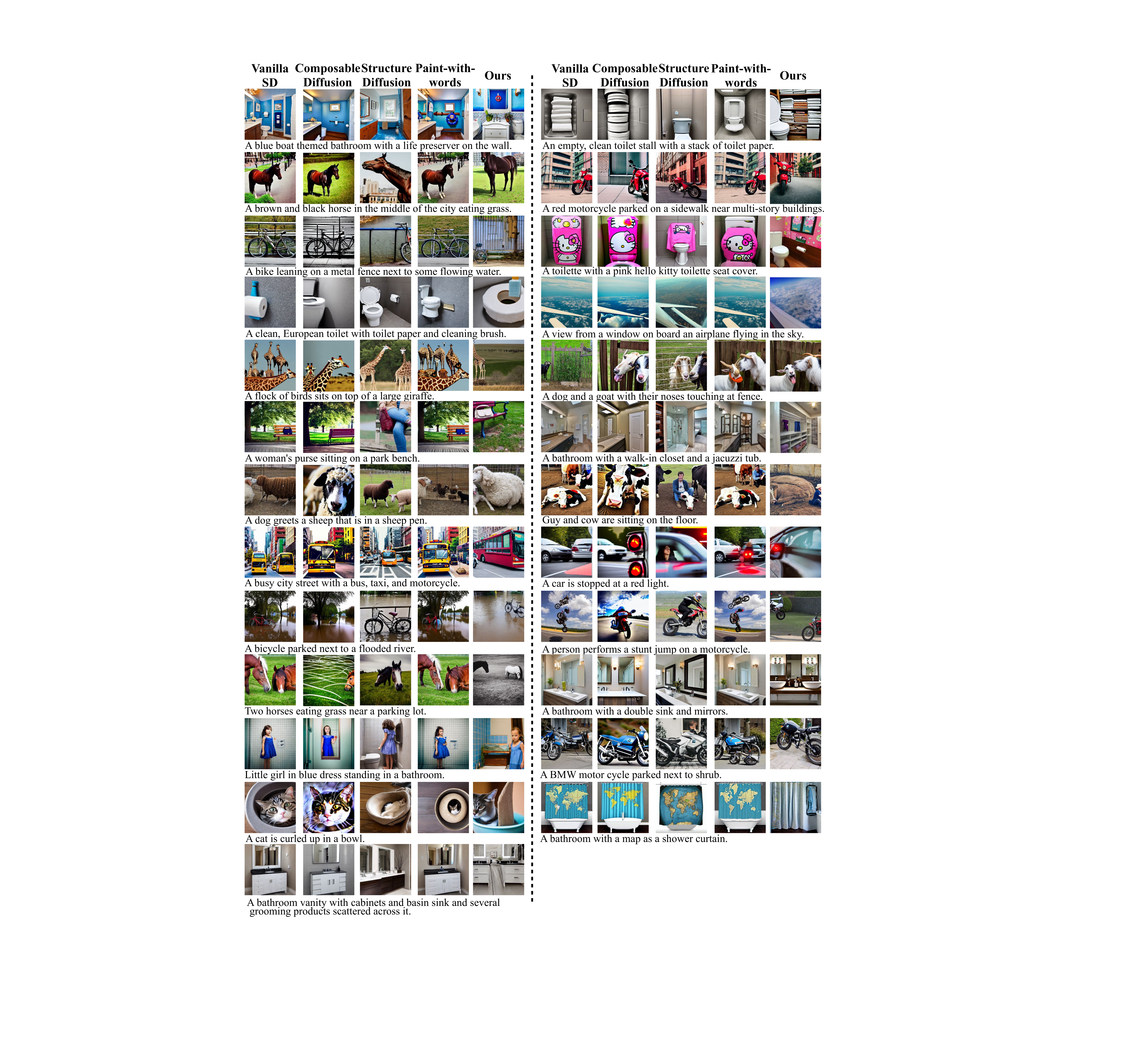}
\caption{Generated images for subjective evaluation on MS-COCO dataset.}
\vspace*{-0.1in}
\label{fig:supp-subjective-mscoco}
\end{figure*}

\end{alphasection}

\end{document}